\documentclass[10pt,journal,compsoc]{IEEEtran}
\ifCLASSOPTIONcompsoc
  \usepackage[nocompress]{cite}
  
\else
  \usepackage{cite}
\fi
\ifCLASSINFOpdf
\else
\fi
\usepackage{amsmath}
\usepackage{algorithm}
\usepackage{algorithmic}
\usepackage{color}
\usepackage{multirow}
\usepackage{amssymb}
\usepackage{caption}
\usepackage{orcidlink}
\usepackage{graphicx}
\usepackage{ragged2e}
\usepackage{booktabs}
\usepackage{graphicx}

\usepackage{academicons}

\begin{document}

\title{Structural and Statistical Texture  Knowledge Distillation and Learning for  Segmentation}

\author{Deyi~Ji$^{\orcidlink{0000-0001-7561-9789}}$,
        Feng~Zhao$^{\orcidlink{0000-0001-6767-8105}}$,
        Hongtao~Lu$^{\orcidlink{0000-0003-2300-3039}}$,  
        Feng~Wu$^{\orcidlink{0000-0001-8451-0881}}$,~\IEEEmembership{Fellow,~IEEE}, 
        and Jieping Ye$^{\orcidlink{0000-0001-8662-5818}}$,~\IEEEmembership{Fellow,~IEEE}
\IEEEcompsocitemizethanks{
\IEEEcompsocthanksitem This work was supported by the Anhui Provincial Natural Science Foundation under Grant 2108085UD12, NSFC (No. 62176155), Shanghai Municipal Science and Technology Major Project, China (2021SHZDZX0102).  We sincerely thank Dr. Lanyun Zhu for the original brilliant work on STLNet. This paper presents STLNet++, which is an improvement based on STLNet. (Corresponding author: Feng Zhao). 

\IEEEcompsocthanksitem Deyi Ji, Feng Zhao, and Feng Wu are with MoE Key Laboratory of Brain-inspired Intelligent Perception and Cognition, University of Science and Technology of China (e-mail: jideyi@mail.ustc.edu.cn, fzhao956@ustc.edu.cn, fengwu@ustc.edu.cn).  \protect 

\IEEEcompsocthanksitem Hongtao Lu is with Department of Computer Science and Engineering, Shanghai Jiao Tong University (e-mail: htlu@sjtu.edu.cn).\protect

\IEEEcompsocthanksitem Jieping Ye is with  Alibaba Group (e-mail: jieping@gmail.com).\protect

\IEEEcompsocthanksitem © 2025 IEEE.  Personal use of this material is permitted.  Permission from IEEE must be obtained for all other uses, in any current or future media, including reprinting/republishing this material for advertising or promotional purposes, creating new collective works, for resale or redistribution to servers or lists, or reuse of any copyrighted component of this work in other works.\protect
} 

}

\IEEEtitleabstractindextext{%
\begin{abstract}
\justifying  
Low-level texture feature/knowledge is also of vital importance for characterizing the local structural pattern and global statistical properties, such as boundary, smoothness, regularity, and color contrast, which may not be well addressed by high-level deep features. In this paper, we aim to re-emphasize the low-level texture information in deep networks for semantic segmentation and related knowledge distillation tasks. To this end, we take full advantage of both structural and statistical texture knowledge and propose a novel Structural and Statistical Texture Knowledge Distillation (SSTKD) framework for semantic segmentation. Specifically, Contourlet Decomposition Module (CDM) is introduced to decompose the low-level features with iterative Laplacian pyramid and directional filter bank to mine the structural texture knowledge, and Texture Intensity Equalization Module (TIEM) is designed to extract and enhance the statistical texture knowledge with the corresponding Quantization Congruence Loss (QDL). Moreover, we propose the Co-occurrence TIEM (C-TIEM) and generic segmentation frameworks, namely STLNet++ and U-SSNet, to enable existing segmentation networks to harvest the structural and statistical texture information more effectively. Extensive experimental results on three segmentation tasks demonstrate the effectiveness of the proposed methods and their state-of-the-art performance on seven popular benchmark datasets, respectively.
\end{abstract}

}

\maketitle

\IEEEdisplaynontitleabstractindextext

%
\IEEEpeerreviewmaketitle

\IEEEraisesectionheading{\section{Introduction}\label{sec:introduction}}

%
%
%
%

\IEEEPARstart{S}{emantic} segmentation, which aims to assign each pixel a unique category label for the input image, is one of the most crucial and challenging tasks in computer vision.
Recently, deep fully convolution network-based methods have achieved remarkable results on semantic segmentation, and extensive methods have been investigated to improve the segmentation accuracy by introducing sophisticated models. 

Despite the promising results that have been achieved, according to the practical application experience in academic and industrial scenarios, there are still two issues worthy of attention. (1) Compact models for  computing on edge devices with limited computation resources. For better segmentation performance, amounts of methods usually utilize large models, introducing tremendous parameters and high inference latency. Since semantic segmentation has shown great potential in many applications like autonomous driving, video surveillance, robot sensing, and so on, how to keep efficient inference speed and high accuracy with high-resolution images is a critical problem. The focus here is knowledge distillation, which is introduced by Hinton et al.  \cite{hinton2015distilling} based on a teacher-student framework, and has received increasing attention in semantic segmentation community.
(2) More fine-grained segmentation characterization on local details. Many practical scene applications, such as lane line segmentation in automatic driving and building segmentation in remote sensing, often require extremely fine-grained local segmentation details to further determine application-level events, for instance, illegal vehicle crossing and building change detection.

For the above two issues, existing methods mainly focus on high-level contextual information/knowledge with large receptive fields, which are appropriate to capture the global context and long-range relation dependencies among pixels. However, they may also result in coarse and inaccurate segmentation results, since the high-level features are usually extracted  with large receptive fields and miss many crucial low-level texture details. Some previous works have been aware of this problem and tried to deliver low-level features to deeper with skip connections (such as U-Net \cite{unet} and DeepLabV3+  \cite{deeplabv3+}) or boundary-aware leaning modules. Representative works of the latter include BFPNet \cite{2020Boundary}, DecoupleSegNet \cite{li2020improving}, and EBLNet \cite{He_2021_ICCV}. 
Yet these methods only focus on the boundary learning and lack a
comprehensive and direct analysis of the low-level texture features.

\begin{figure}[!ht]
  \centering
  \includegraphics[width=1\linewidth]{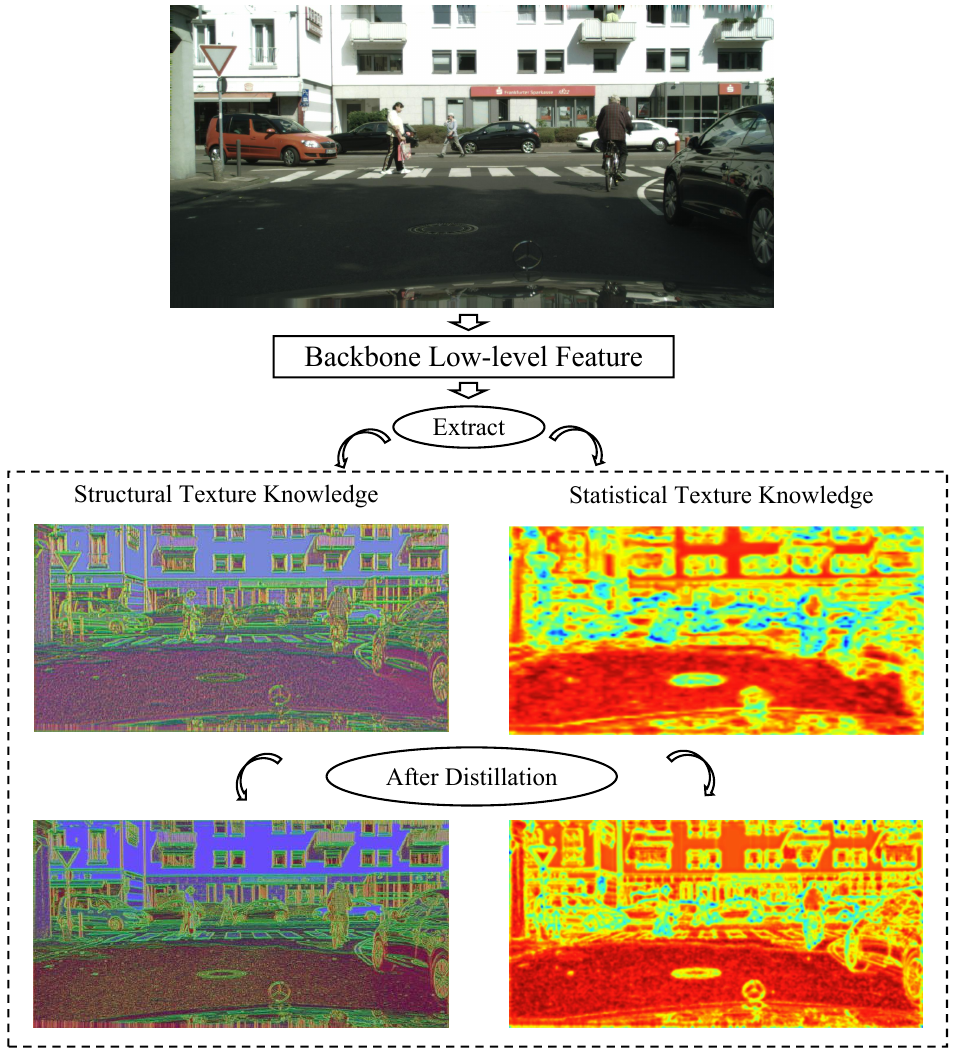}
  \caption{
  An example of the structural and statistical texture knowledge distillation. 
  The original texture is fuzzy and in low-contrast. After distillation, the contour is clearer and the intensity contrast is more equalized, indicating a clear enhancement of both texture types.
  }
  \label{intro}
\end{figure}

In this paper, we concentrate on exploiting unified texture information to boost the performance of both knowledge distillation frameworks and segmentation networks, simultaneously solving both issues mentioned above. According to digital image processing \cite{dip}, texture is a region descriptor that can provide measures for both local structural property and global statistical property of an image. The structural property can also be viewed as spectral domain analysis and often refers to some local patterns, such as boundary, smoothness, and coarseness. In contrast, the statistical property pays more attention to the global distribution analysis, such as histogram of intensity. These properties are essential for pixel-wise segmentation, but are not well addressed in high-level features.

To this end, we first propose a novel Structural and Statistical Texture Knowledge Distillation (SSTKD) framework to effectively distillate two kinds of the texture knowledge from the teacher model to enrich the low-level information for the student model, as shown in Fig. \ref{intro}. More comprehensively,  we propose a Contourlet Decomposition Module (CDM) that decomposes low-level features to mine the structural texture knowledge with iterative Laplacian pyramid and directional filter bank. The contourlet decomposition is a kind of multi-scale geometric analysis tool and can enable the neural network with the ability of geometric transformations, thus it is naturally suitable for describing the structural properties. Meanwhile, we introduce a Texture Intensity Equalization Module (TIEM) to adaptively extract and enhance the statistical knowledge, cooperating with an Anchor-Based Importance Sampler. The TIEM can effectively describe the statistical texture intensity in deep neural networks in a statistical manner, and suppress the noise produced by the amplification effect in near-constant regions during the texture equalization.

Moreover, we design the Co-occurrence TIEM (C-TIEM) and a generic segmentation framework, namely STLNet++, to enable existing segmentation networks to harvest the low-level statistical texture information more effectively. In order to enlarge the scope of STLNet++, we also verify the framework on ultra-high-resolution (UHR) image segmentation task, which has raised increasing demand and interests in recent years. Based on the classical collaborative global-local network \cite{glnet}, a variant of STLNet++, namely U-SSNet, is also presented.

Overall, the main contributions are as follows.
\begin{itemize}
\setlength{\parskip}{0pt} \setlength{\itemsep}{0pt plus 1pt}
    \item We introduce both the structural and statistical texture to knowledge distillation for semantic segmentation, and propose a novel Structural and Statistical Texture Knowledge Distillation (SSTKD) framework to effectively extract and enhance the unified texture knowledge. 
    \item More comprehensively, we present the Contourlet Decomposition Module (CDM) and Texture Intensity Equalization Module (TIEM) to describe the structural and statistical texture, respectively. 
    \item Based on TIEM, we further propose a generic semantic segmentation framework, namely STLNet++. A variant named U-SSNet is also devised for UHR segmentation. 
    \item Extensive experimental results on three segmentation tasks demonstrate the effectiveness of the proposed methods, showing that SSTKD, STLNet++, and U-SSNet achieve state-of-the-art performance on several popular benchmark datasets, respectively. 
\end{itemize}

The remainder of this paper is organized as follows. Sec. \ref{sec:related} reviews the related literature. Secs. \ref{sec:sstkd} and \ref{sec:stlnet++} describe the approaches of SSTKD and STLNet++ (including U-SSNet), respectively. Secs. \ref{sec:exp_kd} and \ref{sec:exp_seg} present the exhaustive experimental results.
Finally, we conclude in Sec. \ref{sec:conclude} and analyze the limitation and broader impact in Sec. \ref{sec:outlook}.

\begin{figure*}[!ht]
    \centering
    \includegraphics[width=1\linewidth]{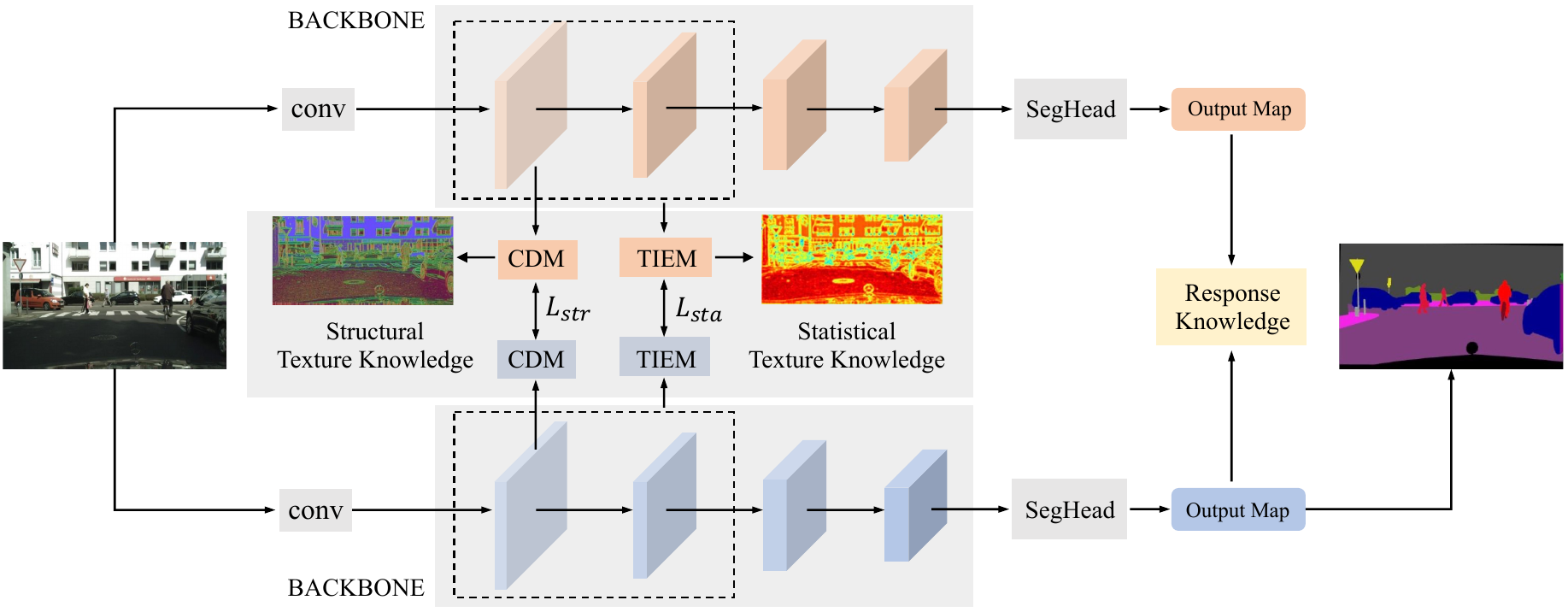}
    \caption{The overview of SSTKD. 
    Apart from the response knowledge, we propose to extract the texture knowledge from low-level features. The corresponding parts of two kinds of texture knowledge are presented in Figs. \ref{cdm}, \ref{lp} and \ref{fig:tiem}, respectively.}
    \label{fig:overview}
\end{figure*}

\section{Literature Review} \label{sec:related}

\subsection{Semantic Segmentation} 

We review the main semantic segmentation methods and directions in the following.

\noindent\textbf{High-Level Context. } Existing works for semantic segmentation mainly focus on contextual information, they try to take advantage of rich contextual information in deep features by enlarging the receptive fields \cite{pspnet, deeplabv3+,feng2018challenges,mscnn}, and using various attention modules \cite{dlpl,zhu2024llafs,zhu2024ibd,zheng2021rethinking,fu2022panoptic,zhu2023continual,zang2025resmatch,zhu2024addressing,liu2021label,chen2024sam2} to harvest the spatial-wise and class-wise dependency. 
During the process, the improvement of network design has also significantly driven the performance, for instance, 
graph modules \cite{ipgn, wang2021learning, cdgcnet, ji2020context,chen2024reasoning3d,chen2024xlstm}.

\noindent\textbf{Low-level Characterization. } Low-level information plays a vital importance on image texture characterization \cite{dlpl,zhu2023learning,homview,chen2023reality3dsketch,zhu2025not,tan2024,tan2025,wangmm,wang2025,zang2025let,wang2023fvp}. Previous works for semantic segmentation usually try to deliver low-level features to deeper layers with skip-connections (e.g., U-Net \cite{unet} and DeepLabV3+  \cite{deeplabv3+}) or boundary-aware leaning modules (e.g., BFPNet \cite{2020Boundary}, DecoupleSegNet \cite{li2020improving}, and EBLNet \cite{He_2021_ICCV}). 
However, they lack a comprehensive and direct analysis of the low-level texture features.

\noindent\textbf{Ultra-High-Resolution (UHR) Image Segmentation. }
Generally, an image with at least 3840×1080 (\~4.1M) pixels reaches the bare minimum bar of ultra-high-resolution (UHR) \cite{glnet,gpwformer,pptformer,xu2025,skysenseo,changenet}. 
Intuitively, working on full UHR images directly will rely on lots of GPU memory, thus leading to the pressing dilemma between memory efficiency and segmentation quality. 
Therefore, recent works \cite{glnet, fctl, isdnet, urur, gpwformer} consider combining their advantages by incorporating the local cropped patches with the corresponding contextual information from downsampled global features.

\noindent\textbf{Texture in Semantic Segmentation. }
From the perspective of digital image processing  \cite{dip}, texture is a kind of descriptor providing measures of properties like smoothness, coarseness, and regularity. Image texture is not only about the local structural patterns, but also global statistical property.  The statistical texture was first introduced to semantic segmentation in \cite{stlnet}. In this work, we focus on the unified texture information including both structural and statistical texture, and propose to improve the statistical characterization.

\subsection{Knowledge Distillation}

In the context of knowledge distillation for dense prediction tasks (e.g., segmentation, detection), the work presented in \cite{detection_survey} provides a comprehensive review that analyzes novel distillation techniques, and stands out as a remarkable survey of the field. This includes an examination of various types of distillation losses and the feature interactions between teacher and student models across extensive object detection applications and datasets. Similarly, in semantic segmentation, early distillation methods yield fundamental results by transferring class probabilities for each pixel from the teacher model to the student model separately. Building upon this foundation, several studies further propose extracting diverse knowledge from a task-specific perspective. For example, SKD \cite{liu2019structured} introduces structured knowledge distillation, which facilitates the transfer of pairwise relations and holistic knowledge through adversarial learning. IFVD \cite{wangintra2020} compels the student model to replicate the intra-class feature variation of the teacher model. CWD \cite{channel_dist} minimizes the Kullback-Leibler divergence between the channel-wise probability maps of the teacher and student networks. CIRKD \cite{CIRKD} leverages the cross-image relational knowledge to transfer global pixel relationships. Distinct from these approaches, we first introduce the texture knowledge into the realm of semantic segmentation, demonstrating an effective framework tailored for this task.

\section{SSTKD: Knowledge Distillation} \label{sec:sstkd}

\subsection{Overview} \label{overview}
The overview of our SSTKD is illustrated in Fig. \ref{fig:overview}. The upper is the teacher module while the lower is the student module. Following \cite{liu2019structured, wangintra2020, channel_dist}, the PSPNet \cite{pspnet} is used for both the teacher (ResNet101) and student (ResNet18). First, we adopt the same basic idea in knowledge distillation to align the response-based knowledge between the teacher and student as previous works \cite{liu2019structured, wangintra2020, channel_dist}. 
Furthermore, we propose to extract two kinds of texture knowledge from the first two layers of the backbone, as the texture information is more reflected on low-level features. For structural texture knowledge, we design a Contourlet Decomposition Module (CDM) that exploits the structural information in the spectral space. For statistical texture knowledge, we introduce a Texture Intensity Equalization Module (TIEM) to adaptively extract the statistical texture intensity histogram with an adaptive importance sampler, then enhance it with a denoised operation and graph reasoning. Finally, we optimize the two types of knowledge between the teacher and student models with two individual texture-related losses.

\subsection{Structural Texture Knowledge Distillation}\label{Structural}

\subsubsection{Structural Texture Extraction}
Traditional filters have inherent advantages for texture representation with different scales and directions in the spectral domain, and structural texture extraction methods are mainly based on geometry  \cite{blostein1989shape,stevens1979surface}, transformation in signal processing  \cite{arof1998circular,kaplan1999extended}, and multi-resolution analysis  \cite{dong2014texture,krishnan2016performance,ccnet}, where methods based on the multi-resolution analysis have gained more attention with higher performance recently. Instead of describing the texture features in the spatial domain, they study the energy distribution in the spectral domain to extract the inherent geometrical structures of texture, and the representative approaches include wavelet transform  \cite{chen2005hybrid,turkoglu2008comparison}, shearlet transform  \cite{dong2014texture,krishnan2016performance}, brushlet transform  \cite{shan2004brushlet}, and contourlet transform  \cite{sha2005unsupervised,hu2006texture,xiangbin2009texture}. Based on the multi-resolution and orientated representations of these transforms, the texture with repeated structure can be extracted effectively. Here, we consider utilizing the contourlet decomposition, a kind of multiscale geometric analysis tool that has substantial advantages in locality and directionality \cite{do2005contourlet, donoho2001can, c-cnn}, and can enhance the ability of geometric transformations in Convolutional Neural Network (CNN).

\begin{figure}[!t]
    \centering
    \includegraphics[width=0.95\linewidth]{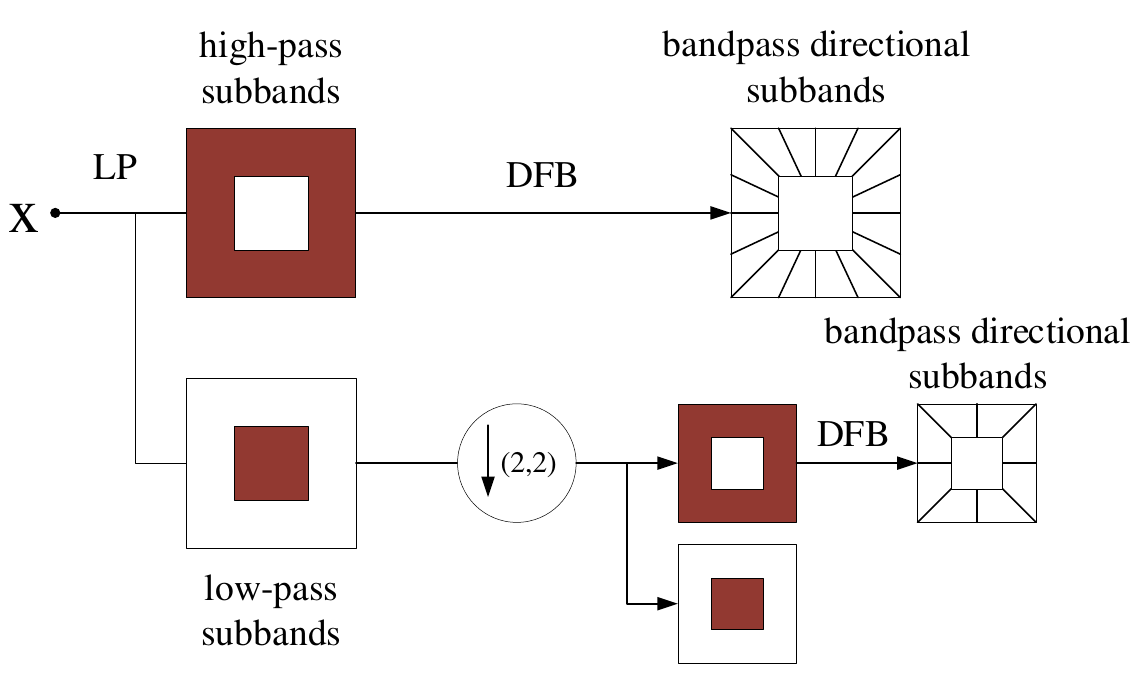}
    \caption{Details of CDM \cite{do2005contourlet, donoho2001can, c-cnn}.}
    \label{cdm}
\end{figure}

\subsubsection{Contourlet Decomposition Module}
\begin{figure}[!t]
    \centering
    \includegraphics[width=0.95\linewidth]{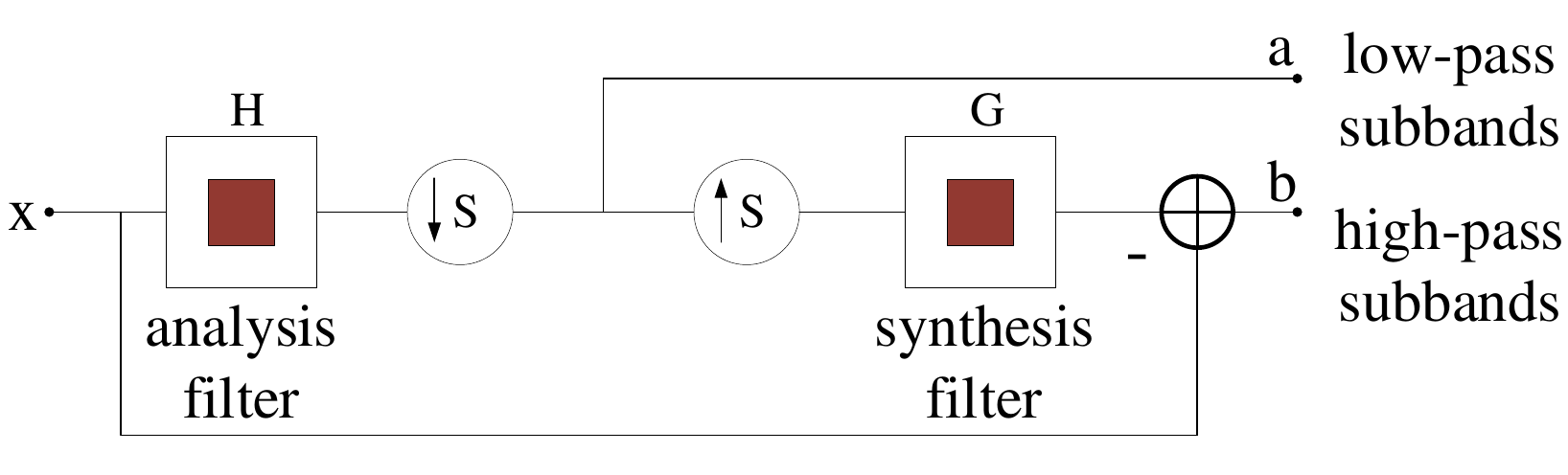}
    \caption{Details of LP decomposition \cite{do2005contourlet, donoho2001can, c-cnn}.}
    \label{lp}
\end{figure}

Based on these advantages, we introduce a Contourlet Decomposition Module (CDM) to mine the texture knowledge in the spectral space.
As shown in Fig. \ref{cdm}, it adopts a Laplacian Pyramid (LP)  \cite{burt1983laplacian} and a directional filter bank (DFB)  \cite{bamberger1992filter} iteratively on the low-pass image.
The LP aims to obtain multi-scale decomposition. As illustrated in Fig. \ref{lp}, given the input feature $x$, a low-pass analysis filter $H$ and a sampling matrix $S$ are used to generate downsampled low-pass subbands. Then, the high-pass subbands are obtained by the difference between the original $x$ and the intermediate result, which is computed by a sampling matrix $S$ and a low-pass synthesis filter $G$. Following that, DFB is utilized to reconstruct the original signal with a minimum sample representation, which is generated by $m$-level binary tree decomposition in the two-dimensional frequency domain, resulting in $2^m$ directional subbands. For example, the frequency domain is divided into $2^3=8$ directional subbands when $m=3$, and the subbands $0$-$3$ and $4$-$7$ correspond to the vertical and horizontal details, respectively.
Finally, the output of the contourlet decomposition in level $n$ can be described by the following equations:
\begin{equation}
    \begin{aligned}
        F_{l,n+1},F_{h,n+1}&=LP(F_{l,n})\downarrow p \\
        F_{bds,n+1}&=DFB(F_{h,n+1}) ~~~ n \in [1, m],
    \end{aligned}
\end{equation}
where symbol $\downarrow$ is the downsampling operator and $p$ denotes the interlaced downsampling factor. The subscripts $l$ and $h$ represent the low-pass and high-pass components respectively, $bds$ indicates the bandpass directional subbands.

For a richer expression, we stack multiple contourlet decomposition layers iteratively in the CDM.
In this way, abundant bandpass directional features are obtained via the contourlet decomposition from the low-level features, which are used as the structural texture knowledge $F^{str}$ for distillation. 
We apply the CDM to the teacher and student networks respectively and use the conventional Mean Squared (L2) loss to formulate the texture distillation loss:
\begin{equation}
        L_{str}(\mathcal{S})=\frac{1}{(W \times H)}\sum_{i\in R}{(F^{str;\mathcal{T}}_{i}-F^{str;\mathcal{S}}_{i})^2},
\end{equation}
where $F^{str;\mathcal{T}}_{i}$ and $F^{str;\mathcal{S}}_{i}$ denote the $i$th pixel in texture features produced from the teacher network $\mathcal{T}$ and the student network $\mathcal{S}$ separately, and $i \in R=W \times H$ that represents the feature size.

\begin{figure}
    \centering
    \includegraphics[width=0.8\linewidth]{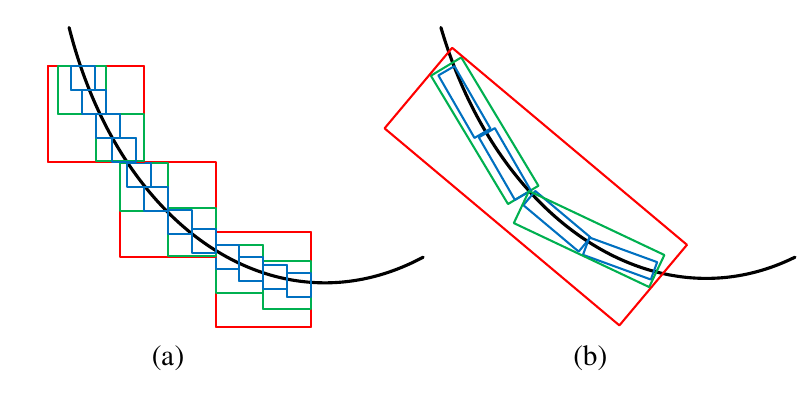}
    \caption{(a) The wavelet possesses square supports, mainly suitable for capturing point discontinuities. (b) The contourlet is able to capture linear segments of contours with fewer coefficients \cite{do2005contourlet, donoho2001can, c-cnn}.}
    \label{discussion_structural}
\end{figure}

\subsubsection{Discussion of Structural Texture Extraction Methods} \label{sec_discuss_structural}

As one of the signal representation tools, contourlet is similar to the edge in an image. It can make full use of geometric regularity to obtain the sparse representation.
Previous works apply fast Fourier transform (FFT) or wavelet decomposition to capture the texture information. For example, \cite{donoho2001can} involves FFT based on the shallow features, with high-frequency features used as texture. However, due to the lack of more precise design, the effectiveness of this approach is often limited in practice. Regarding the wavelet decomposition, as shown in Fig. \ref{discussion_structural}(a), there are largely inefficient decompositions as points in the wavelet decomposition. In contrast, Fig. \ref{discussion_structural}(b) shows that the contourlet decomposition can describe a smooth contour as line segments with much fewer coefficients. Thus, the representation with the contourlet decomposition is sparser, and has advantages of directionality, locality, and the bandpass property, which is consistent with the main characteristics of the human visual system (HVS).
In conclusion, contourlet has elongated support intervals at various scales and directions with more sparse representation, which can better describe the different directional contours in texture knowledge.

\subsection{Statistical Texture Knowledge Distillation}\label{Statistical}

The statistical texture is usually of a wide variety and a continuous distribution in the spectral domain, which is difficult to be directly extracted and optimized in deep neural networks. Towards this goal, inspired by the theory of  Histogram Equalization (HE) \cite{ahe}, we propose the Texture Intensity Equalization Module (TIEM) to describe and enhance the texture intensities in a statistical manner in deep neural networks. TIEM consists of four steps including Quantization, Counting, Denoising, and Equalization. It aims to quantize the input feature into multiple levels along with a denoising process, then counts the number of features belonging to each level to construct the intensity histogram, followed by a graph module to perform intensity equalization. 
Finally, we illustrate the distillation process and propose the Texture Distillation Loss (TDL).

\subsubsection{Anchor-Based Importance Sampler} \label{anchor_sampler}
Extracting the statistical texture of the whole input feature is straightforward, but it lacks the attention to discriminative regions and only works well when the feature intensities of pixels are near constantly distributed. However, real-world scenes are usually offended with chaotic conditions and the pixel intensities tend to follow a heavily imbalanced distribution.  
Thus, we consider using an importance sampling method to mine the hard-to-classify areas, where the feature intensity distribution is diverse and the statistical texture is rich. Based on \cite{pointrend},
we propose an Anchor-Based Importance Sampler. It is designed to bias selection towards most uncertain regions, while retaining some degree of uniform coverage, by the following steps. \textit{\lowercase\expandafter{(\romannumeral1)} Over Generation}: Aiming at sampling $M$ points, to guarantee the variety and recall, we over-generate candidate points by randomly selecting the $kM (k>1)$ points with a uniform distribution. \textit{\lowercase\expandafter{(\romannumeral2)} Importance Sampling}: Among the $kM$ points, we intend to choose the most uncertain $\beta M (\beta \in [0, 1])$ ones with an anchor-based  importance sampling strategy. \textit{\lowercase\expandafter{(\romannumeral3)} Coverage}: To balance the distribution, we select the remaining $(1-\beta) M$ points from the rest of the points with a uniform distribution.

In \textit{importance sampling}, for each sample $s \in kM$, we set 9 anchors with 3 scales and 3 aspect ratios at the location. In this way, we generate 9 region proposals $\{\mathbf{r}_i\}(i\in[1,9])$ for each $s$, and its sample probability can be calculated by, 

\begin{equation}
\begin{aligned}
    p_{s} = \sum_{i} std(\mathbf{r}_{i}), 
\end{aligned}
\end{equation}
\noindent where $std(\cdot)$ means the variance function. It shows that  $s_i$ with larger region variance will be more likely sampled, as the region intensity distribution is diverse and contains richer statistical texture that needs to be enhanced. For each sampled $s$, we randomly select one region $\mathbf{r}$ from $\{\mathbf{r}_i\}$ for the subsequent process. Note that the student utilizes the same importance sampling results as the teacher to select region proposals. Of course, since the sizes of feature maps of the student are often smaller than those of the teacher, its selected regions should also be proportionally scaled down.

\begin{figure*}[!ht]
    \centering
    \includegraphics[width=1\linewidth]{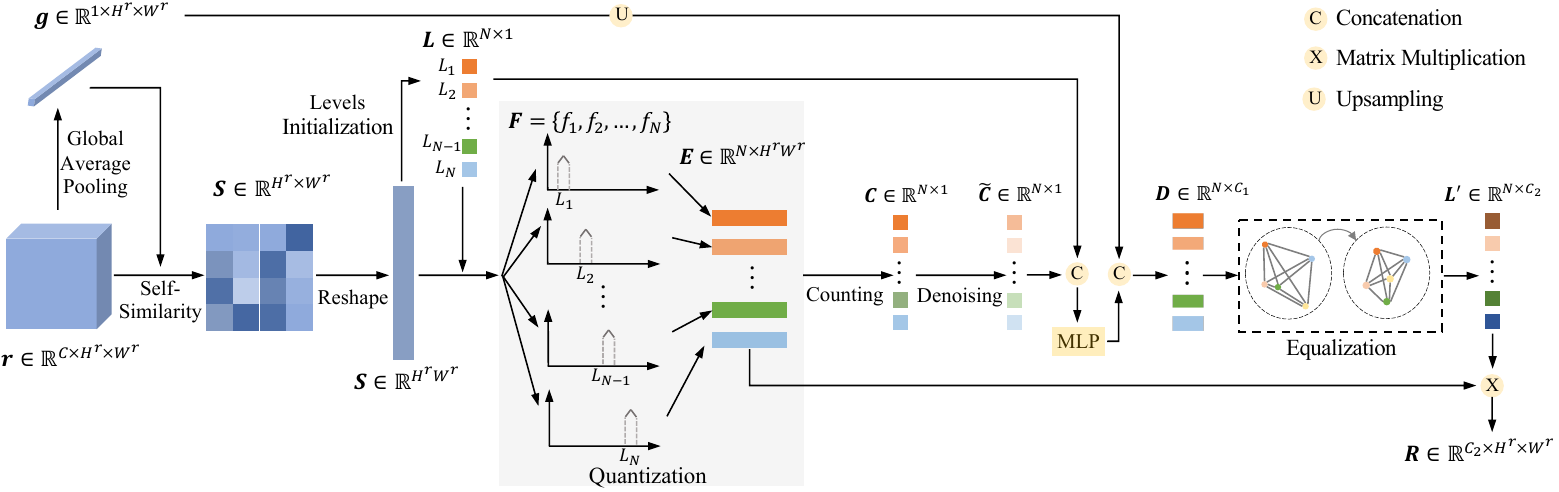}
    \caption{The detailed process of TIEM. }
    \label{fig:tiem}
\end{figure*}

\subsubsection{TIEM: Quantization Step}

\begin{figure}
    \centering
    \includegraphics[width=0.95\linewidth]{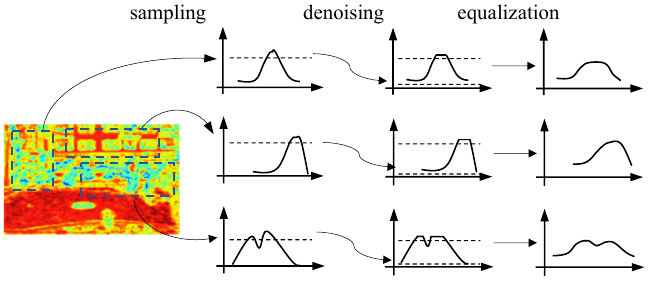}
    \caption{An intuitive description about the process of TIEM.}
    \label{TIEM_detail}
\end{figure}

As shown in Fig. \ref{fig:tiem} and Fig. \ref{TIEM_detail}, given a sampled region $\mathbf{r} \in\mathbb{R}^{C\times H^r \times H^r}$ from the input feature map  $\mathbf{A} \in \mathbb{R}^{C\times H \times W}$, the self-similarity matrix $\mathbf{S}\in \mathbb{R}^{1\times H^r \times W^r}$ is first calculated. Specifically, each position $\mathbf{S}_{i,j}$ $(i \in [1, H^r], j \in [1, W^r])$ of $\mathbf{S}$ is denoted as,
\begin{equation}
    \mathbf{S}_{i,j} = \frac{\mathbf{g} \cdot \mathbf{A}_{i,j}}{\lVert \mathbf{g}\lVert_{2} \cdot \lVert \mathbf{A}_{i,j}\lVert_{2}},
\end{equation}
\noindent where $\mathbf{g} \in \mathbb{R}^{C\times1\times1}$ is the global averaged feature of $\mathbf{r}$ calculated by a global average pooling operator. $\mathbf{S}$ is then reshaped to $\mathbb{R}^{H^rW^r}$ and quantized into $N$ levels: $\mathbf{L} = \{\mathbf{L}_{1}, \mathbf{L}_{2}, ..., ..., \mathbf{L}_{N}\}$, which is obtained by equally dividing the $N$ points between the minimum and maximum values of $\mathbf{S}$. The $n$th ($n \in [1, N]$) level $ \mathbf{L}_{n}$ is calculated by,
\begin{equation}
    \mathbf{L}_{n} = \frac{max\left(\mathbf{S}\right) - min\left(\mathbf{S}\right)}{N} \cdot n + min(\mathbf{S})
    \label{l}.
\end{equation}

\noindent For each spatial pixel $\mathbf{S}_{i} \in \mathbb{R}(i\in[1,H^rW^r])$, we quantize it to a quantization encoding vector $\mathbf{E}_i \in \mathbb{R}^{N}(i\in[1,H^rW^r])$, thus $\mathbf{S}$ will be finally quantized to a quantization encoding matrix $\mathbf{E} \in \mathbb{R}^{N \times H^rW^r}$. More comprehensively, $\mathbf{S}_{i}$ is quantized with $N$ functions $F=\{f_1, f_2, ...,f_N\}$, where each $f_n$ produces $\mathbf{E}_{i,n}$ by,

\begin{equation}
    \resizebox{0.9\hsize}{!}{$\mathbf{E}_{i,n} = \left\{
        \begin{array}{lcc}
        1-|\mathbf{L}_{n}-\mathbf{S}_{i}|, ~~~~  if  ~~  -\frac{0.5}{N}\le \mathbf{L}_{n}-\mathbf{S}_{i}<\frac{0.5}{N}\\
        0,  ~~~~~~~~~~~~~~~~~~~~~~~ else  
    \end{array}.
    \right.$}
\end{equation}

\noindent Finally, the $N$ results of $F$ are concatenated to get $\mathbf {E}_{i}$. By this way, the index of maximum value of $\mathbf{E}_{i}$ can reflect the quantization level of $\mathbf{S}_{i}$. 
Note that, in contrast to traditional argmax operation or one-hot encoding mechanism with binarization, we apply a smoother way for quantization encoding so that the gradient vanishing problem can be avoided during the process of back propagation.

\subsubsection{TIEM: Counting Step}
Given the quantization encoding matrix $\mathbf{E}$, the two-dimensional quantization counting map $\mathbf{C} \in \mathbb{R}^{N \times 1}$ is then calculated, where the first dimension represents each quantization level and the second dimension represents the corresponding normalized counting number. Formally, $\mathbf{C}$ is calculated by,
\begin{equation}
    \mathbf{C} = \frac{\sum_{i=0}^{HW}{\mathbf{E}_{i}^{n}}}{\sum_{n=1}^{N}\sum_{i=1}^{HW}{\mathbf{E}_{i}^{n}}}.
\end{equation}
\noindent In the case where TIEM is analogous to Histogram Equalization (HE), $\mathbf{C}$ plays a role of histogram here.

\subsubsection{TIEM: Denoising Step}

Before applying equalization to $\mathbf{C}$, a Denoising Step is first applied. According to the theory of Contrast-Limited  Histogram Equalization (CLHE)  \cite{ahe}, ordinary HE tends to over-amplify the contrast in near-constant regions of the image, since the histogram in such regions is highly concentrated. As a result, HE may cause noise to be amplified in near-constant regions. CLHE is a variant of HE in which the contrast amplification is limited, so as to reduce this problem of noise amplification  \cite{ahe,wiki_ahe}, in concrete, by clipping the part of the histogram that exceeds a predefined value and then redistributing it equally among all histogram bins.
Heuristically, our quantization process may also suffer from the quantization noise produced by the over-amplifying effect in near-constant areas. Correspondingly, we exploit an intensity-limited Denoising Step after the Quantization Step to constrain the intensity peaks of $\mathbf{C}$ with a given threshold $\theta$, then redistribute the sum of extra peaks $\mathbf{C}_{e}$ equally among all the quantization levels.
Finally, the denoised quantization counting map $\mathbf{\widetilde{C}}$ can be calculated by,

\begin{equation}
    \mathbf{C}_{e} = \sum_{n}^{N} [\mathop{max}(\mathbf{C}_n - \theta \cdot \mathop{max}\limits_{N}(\mathbf{C}), 0)],
\end{equation}

\begin{equation}
\centering
    \resizebox{.9\hsize}{!}{$\mathbf{\widetilde{C}}_{n} = \left\{
        \begin{array}{lcc}
        \theta \cdot \mathop{max}\limits_{N}(\mathbf{C}) + \frac{\mathbf{C}_{e}} {N},  ~~~~ if ~~ \mathbf{C}_{n} > \theta \cdot \mathop{max}\limits_{N}(\mathbf{C})  \\
        \mathbf{C}_{n} + \frac{\mathbf{C}_{e}} {N},  ~~~~~~~~~~~~~~~~ else  
    \end{array}
    \right.,$}
    \label{Edfunc}
\end{equation}

\noindent where $n \in [1, N]$, $max(\cdot)$ is the maximum function. 

Then, $\mathbf{\widetilde{C}}$ is able to reflect the relative statistics of the input feature map, as the counted object is the distance from the average feature. To further obtain absolute statistical information, we encode the quantization levels $\mathbf{L}$ and global average feature $\mathbf{g}$ into $\mathbf{\widetilde{C}}$, and produce the statistical feature $\mathbf{D}\in \mathbb{R}^{N\times C_1}$ by,
\begin{equation}
    \begin{split}
        \mathbf{D} = Cat(MLP(Cat(\mathbf{L}, \mathbf{\widetilde{C}})), \mathbf{g}),
    \end{split}
\end{equation}
where $Cat(\cdot)$ denotes the concatenation operation, $\mathbf{g}$ is first upsampled to $\mathbb{R}^{N\times C}$, and $MLP$ aims to increase the dimension for $Cat(\mathbf{L}, \mathbf{\widetilde{C}})$. It contains two layers, in which the first one is followed by a Leaky ReLU to enhance the nonlinearity. Note that $\mathbf{D}$ is the denoised variant of $\mathbf{C}$, thus also playing the role of histogram.

\subsubsection{TIEM: Equalization Step}

Low-level features extracted from shallow layers of the backbone network are often of low quality, especially with low contrast, causing the texture details 
to be ambiguous, which has negative impacts on the extraction and utilization of low-level information. Therefore, an Equalization Step is finally applied to the statistical feature $\mathbf{D}$ to
enhance the texture details in low-level features, so that it is easier to capture texture-related information in the following steps. 

The way we enhance texture is analogous to  Histogram Equalization (HE), a classical method for image quality enhancement. Specifically, this method first produces a histogram whose horizontal axis and vertical axis represent each gray level and its counting value, respectively. We denote these two axes as two feature vectors $\mathbf{G}$ and $\mathbf{F}$. HE aims to reconstruct levels to $\mathbf{G}'$ using the statistical information containing in $\mathbf{F}$. Each level 
$\mathbf{G}_{n}$ is transformed into $\mathbf{G}'_{n}$ by, 
\begin{equation} \label{gcn}
    \mathbf{G}'_{n} = \frac{(N-1)\sum_{i=0}^{n}\mathbf{F}_{n}}{\sum_{i=0}^{N}\mathbf{F}_{n}},
\end{equation}
where $N$ is the total number of gray levels.

In the first three steps of TIEM, we obtain the quantization encoding matrix $\mathbf{E}\in \mathbb{R}^{N\times HW}$ and statistics feature $\mathbf{D}\in \mathbb{R}^{N \times C_{1}}$, where $\mathbf{D}$ plays the role of histogram. After that, we aim to obtain the reconstructed quantization levels $\mathbf{L}'$ with $\mathbf{D}$. Inspired by Equation \ref{gcn}, each new level should be obtained by perceiving statistical information of all original levels, which can be treated as a graph. To this end, we build a graph to propagate information from all levels. The statistical 
feature of each quantization level is defined as a node. In the traditional HE algorithm, the adjacent matrix is a manually defined diagonal matrix, and we extend it to a learned one as follows.
\begin{equation}
    \mathbf{X} = Softmax(\phi_{1}(\mathbf{D}) \cdot \phi_{2}(\mathbf{D})^{\top}),
\end{equation}
where $\phi_{1}$ and $\phi_{2}$ refer to two different $1\times 1$ convolutions, $\top$ is the matrix transpose operation, and Softmax performed on the first dimension works as a nonlinear normalization function.
Then, we update each node by fusing features from all the other nodes, getting the reconstructed quantization levels $\mathbf{L}'\in \mathbb{R}^{N \times C_{2}}$ by,
\begin{equation}
    \mathbf{L}' = \mathbf{X} \cdot \phi_{3}\left(\mathbf{D}\right),
\end{equation}
where $\phi_{3}$ refers to another $1\times 1$ convolution. 

Next, we assign the reconstructed levels $\mathbf{L}'$ to each pixel to get the final output $\mathbf{R}$ with the quantization encoding map $\mathbf{E}\in \mathbb{R}^{N\times HW}$, since $\mathbf{E}$ can reflect the original quantization level of each pixel. $\mathbf{R}$ is obtained by,
\begin{equation}
    \mathbf{R} = \mathbf{L}'^{\top} \cdot \mathbf{E}.
\end{equation}
Finally, $\mathbf{R}$ is reshaped to $\mathbf{R}^{C_{2}\times H\times W}$.

\subsubsection{Quantization Congruence Loss} 

We propose the Quantization Congruence Loss (QCL) to explicitly promote the congruence on both statistical feature $\mathbf{D}$ and reconstructed quantization levels $\mathbf{L}^{'}$ between teacher and student. First, for each region $\mathbf{r}$, similar as the structural texture distillation, we apply $L_2$ loss to formulate the statistical feature distillation:
\begin{equation}
        L_{D}(\mathcal{S})=\frac{1}{(W^r \times H^r)}\sum_{i\in W^rH^r}{(\mathbf{D}_{i}^{\mathcal{T}}-\mathbf{D}_{i}^{\mathcal{S}})^2}.
\end{equation}
\noindent However, it is not appropriate to apply the $L_2$ loss to the distillation of quantization levels for two reasons: (1) the generation of levels is strongly dependent on the intensities of feature map, so the levels in student and teacher are inconsistent in scale; (2) even if scale normalization is used, pushing student to align the values of levels to teacher exactly will lead to network instability, thus too hard to optimize. On the contrary, we actually expect to promote student to lean the distribution among the levels, which is the key we need to characterize. An easy-to-think and intuitive way is to calculate the correlation between any two levels, but the computational complexity of such an approach increases squarely as the number of levels grows, which is not applicable. Instead, we express the characterization as the distance between each level and the overall, with a linear complexity. Here, Mahalanobis distance is used to take into account the correlations of the levels set, since it is unitless and scale-invariant.
Formally, given the levels $\mathbf{L}^{'}$, without loss of generality, correlation of each level $\mathbf{L}^{'}_{n}$ with the overall can be expected as,
\begin{equation}
    Corr(\mathbf{L}^{'}_n, \mathbf{L}^{'})= \sqrt{(\mathbf{L}^{'}_n - \mu_0)P^{-1}(\mathbf{L}^{'}_n - \mu_0)},
\end{equation}
\noindent where $\mu_0$ is the average of $\mathbf{L}^{'}$, and $P$ is the positive-definite covariance matrix. Thus, the QDL is expressed as, 
\begin{equation}
\begin{aligned}
    L_{qdl}(\mathcal{S}) = L_D(\mathcal{S}) + & (\sum_{n^{\mathcal{S}} \in N^{\mathcal{S}}} Corr(\mathbf{L}^{';\mathcal{S}}_n, \mathbf{L}^{';\mathcal{T}}) - \\
    & \sum_{n^{\mathcal{T}} \in N^{\mathcal{T}}} Corr(\mathbf{L}^{';\mathcal{T}}_n, \mathbf{L}^{';\mathcal{T}})),
\end{aligned}
\end{equation}
\noindent which is served as the statistical texture distillation loss $L_{sta}(\mathcal{S})$.

{\subsubsection{Discussion of Statistical Texture Extraction Methods} \label{sec_discuss_statistical}

In this paper, we introduce the Texture Intensity Enhancement Model (TIEM) to describe and enhance the texture intensities in deep neural networks in a statistical manner. Previous works have utilized the Gray-level Co-occurrence Matrix (GLCM) to capture important statistical textural information \cite{glcm1, glcm2, glcm3, glcm4}. The primary principle behind GLCM involves executing operations based on second-order statistics derived from the image. When integrating GLCM into Convolutional Neural Networks (CNNs), the process begins with gray-level scaling to constrain the image values using a histogram-based adaptive method \cite{glcm1, glcm2}. Subsequently, the co-occurrence matrix is generated to represent the final statistical texture features. However, GLCM-based methods often necessitate hand-crafted computations and rely on shallow networks to extract low-dimensional texture features, resulting in limited performance.

In contrast, our proposed TIEM adaptively extracts high-dimensional statistical texture features and enhances them through a denoising operation and graph reasoning, based on Contrast-Limited Histogram Equalization (CLHE) theory. Consequently, the statistical texture features obtained through TIEM exhibit greater strength and robustness compared to traditional methods.

\subsection{Overall Distillation Loss} \label{Optimization}

Following the common practice and previous knowledge distillation works for semantic segmentation  \cite{liu2019structured, wangintra2020, structure_dense, channel_dist}, we also add the fundamental response-based distillation loss $L_{re}$ and adversarial loss $L_{adv}$ for stable gradient descent optimization.
The response-based distillation loss can add a strong congruent constraint on predictions, and is usually expressed as the Kulllback-Leiber (KL) divergence between the class probabilities yielded from the teacher and student models. Formally, $L_{re}$ is defined as follows.
\begin{equation}
    L_{re} = \frac{1}{(W^{re} \times H^{re})} \sum_{i\in R} KL(P_i^{re;\mathcal{T}} || P_i^{re;\mathcal{S}}),
\end{equation}
where $P_i^{re;\mathcal{T}}$ and $P_i^{re;\mathcal{S}}$ denote the class probabilities of $i$-th pixel produced by the teacher and student models respectively, and $i \in R=W^{re} \times H^{re}$ represents the output size. The adversarial training aims to formulate the holistic distillation problem \cite{liu2019structured} and is denoted by,
\begin{equation}
    L_{adv} = \mathbb{E}_{\mathcal{S} \sim p(\mathcal{S})}[D(\mathcal{S}|\mathcal{I})],
\end{equation}
where $\mathbb{E}(\cdot)$ is the expectation operator, $D(\cdot)$ is the discriminator, and $\mathcal{I}$ as well as $\mathcal{S}$ are the input image and corresponding segmentation map, respectively.

Therefore,  for the overall optimization, the whole objective function consists of a conventional cross-entropy loss $L_{seg}$  for semantic segmentation and the above-mentioned distillation loss:
\begin{equation}
\begin{aligned}
    L = L_{seg} + \lambda_1(L_{str} + L_{sta}) + \lambda_2L_{re} - \lambda_3 L_{adv},
\end{aligned}
\end{equation}
where $\lambda_1, \lambda_2, and \lambda_3$ are set to 0.9, 3, and 0.01, respectively.

\section{STLNet++ and U-SSNet: Semantic Segmentation} \label{sec:stlnet++}

We propose a generic segmentation framework, namely STLNet++, to take advantage of the statistical texture in deep convolution networks for semantic segmentation. As shown in Fig. \ref{stlnet++}, we first apply TIEM to the low-level features to enhance the texture information. Then, C-TIEM is proposed for deep co-occurrence texture representation, which is then fused with the classical contextual feature for final prediction. The formulation of C-TIEM is inspired by Gray-level Co-occurrence Matrix (GLCM), which is defined on an image as the distribution of co-occurring pixel values (grayscale values or colors) at a given offset. It is often used as an approach to texture analysis, since texture is highly correlated with the statistical information about spatial relationships between pixels.

\subsection{C-TIEM: Dense Dilated Co-occurrence TIEM}

Although TIEM can capture the distribution of pixel intensity, it neglects the  spatial location association among pixels. Before applying TIEM to the segmentation networks, we first enable it to express the dense quantization co-occurrence in spatial locations of the input feature map, namely C-TIEM, inspired by the theory of Gray-level Co-occurrence Matrix  \cite{dip}. Similarly, C-TIEM consists of Quantization, Counting, Denoising, and Adaptation steps, and is directly applied to the whole input feature map.

\subsubsection{C-TIEM: Quantization Step}

Following the similar procedure as TIEM, the quantization encoding map $\mathbf{E} \in \mathbb{R}^{H \times W \times N \times 1}$ and quantization levels $\mathbf{L} \in \mathbb{R}^{N \times 1}$ are first obtained with the input $\mathbf{A}\in \mathbb{R}^{C \times H \times W}$. Multiple dilation steps (\{1,3,5\}) are set to generate the dense quantization encoding map. For each step $s$, the quantization co-occurrence of spatial pixel $\mathbf{A}_{i,j}$ and $\mathbf{A}_{i, j+s}$ is calculated by,

\begin{equation}
    \mathbf{\hat{E}}_{i, j}^s = \mathbf{E}_{i, j} \cdot \mathbf{E}_{i, j+s}^{\top},
\end{equation}

\noindent where $\top$ is the matrix transposition operator. It is found that $\mathbf{\hat{E}}^s \in \mathbb{R}^{H \times W \times N \times N}$ and $\mathbf{\hat{E}}^s_{m, n, i, j}  \neq 0$ only when $\mathbf{A}_{i, j}$ is quantized to $\mathbf{L}_m$ and $\mathbf{A}_{i, j+s}$ is quantized to $\mathbf{L}_n$, where $m, n \in [1, N], i \in [1, W], j \in [1, H]$.

\subsubsection{C-TIEM: Counting Step}

Given  $\mathbf{\hat{E}}^s$, we generate a 3-dimensional map $\mathbf{C}^s \in \mathbb{R}^{N \times N \times 1}$, where the first two dimensions represent each possible quantization co-occurrence with step $s$, and the third dimension represents the corresponding normalized counting number. $\mathbf{C}^s$ is denoted as,
\begin{equation}
\begin{split}
    &\mathbf{C}^s =  \frac{\sum_{i=1}^{H}\sum_{j=1}^{W}{\mathbf{\hat{E}}^s_{m, n, i, j}}}{\sum_{m=1}^{N}\sum_{n=1}^{N}\sum_{i=1}^{H}\sum_{j=1}^{W}{\mathbf{\hat{E}}_{m, n, i, j}^s}}.
\end{split}
\end{equation}

\begin{figure}[!ht]
    \centering
    \includegraphics[width=0.95\linewidth]{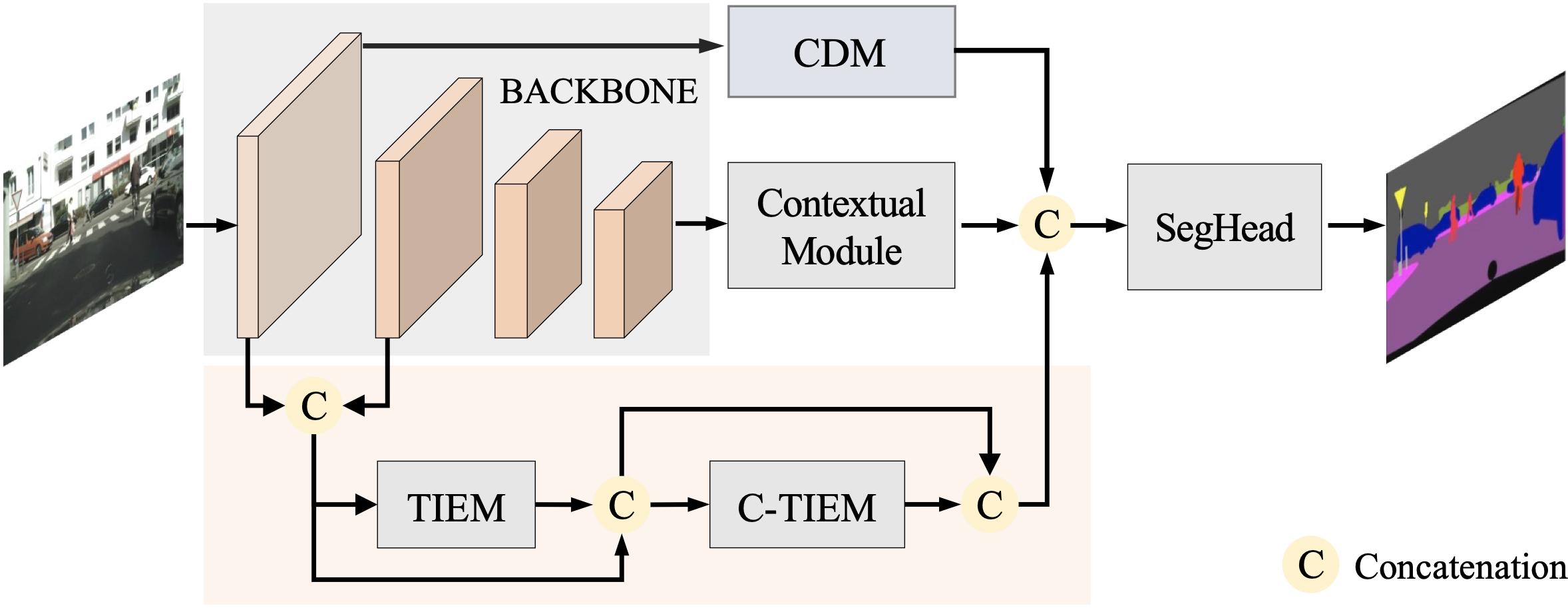}
    \caption{An overview of our proposed STLNet++. }
    \label{stlnet++}
\end{figure}

\subsubsection{C-TIEM: Denoising Step}

The intensity-limited denoising strategy is also performed on $\mathbf{C}^s$ to produce the denoised quantization counting map $\mathbf{\widetilde{C}}^s \in \mathbb{R}^{N \times N \times 1}$ by, 

\begin{equation}
    \mathbf{C}_{e}^s = \sum_{m}^{N} \sum_{n}^{N} [\mathop{max}(\mathbf{C}^s - \theta \cdot \mathop{max}_{N,N}(\mathbf{C}^s), 0)],
\end{equation}

\begin{equation}
    \resizebox{.87\hsize}{!}{$\mathbf{\widetilde{C}}^s_{m,n} = \left\{
        \begin{array}{lcc}
        \theta \cdot \mathop{max}\limits_{N,N}(\mathbf{C}^s) + \frac{\mathbf{C}_e^s}{N^2}  ~~~ if ~~    \mathbf{C}_{m,n}^s > \theta \cdot \mathop{max}\limits_{N,N}(\mathbf{C}^s) \\
        \mathbf{C}^s_{m,n} + \frac{\mathbf{C}_e^s}{N^2}  ~~~~~~~~~~~~~~~~~~~~~~~~~~~~   else
    \end{array}
    \right..$}
    \label{Edfunc}
\end{equation}
\noindent Finally, the co-occurrence statistical map $\mathbf{D} \in \mathbf{R}^{C_2 \times N \times N}$ is denoted as,
\begin{equation}
\begin{aligned}
    &\mathbf{D} = \mathop{Cat}\limits_{s}(\mathop{Cat}(MLP(Cat(\hat{\mathbf{L}}, \mathbf{\widetilde{C}}^s)), \mathbf{g})),\\
    &\resizebox{.87\hsize}{!}{$\hat{\mathbf{L}} \in \mathbb{R}^{2\times N\times N}, \hat{\mathbf{L}}_{m, n} = [\mathbf{L}_{m}, \mathbf{L}_{n}], m\in[1, N], n\in[1,N],$}
\end{aligned}
\end{equation}
\noindent where $\hat{\mathbf{L}}$ denotes all possible quantization level pairs of adjacent pixels.

\begin{figure*}[!ht]
    \centering
    \includegraphics[width=0.85\linewidth]{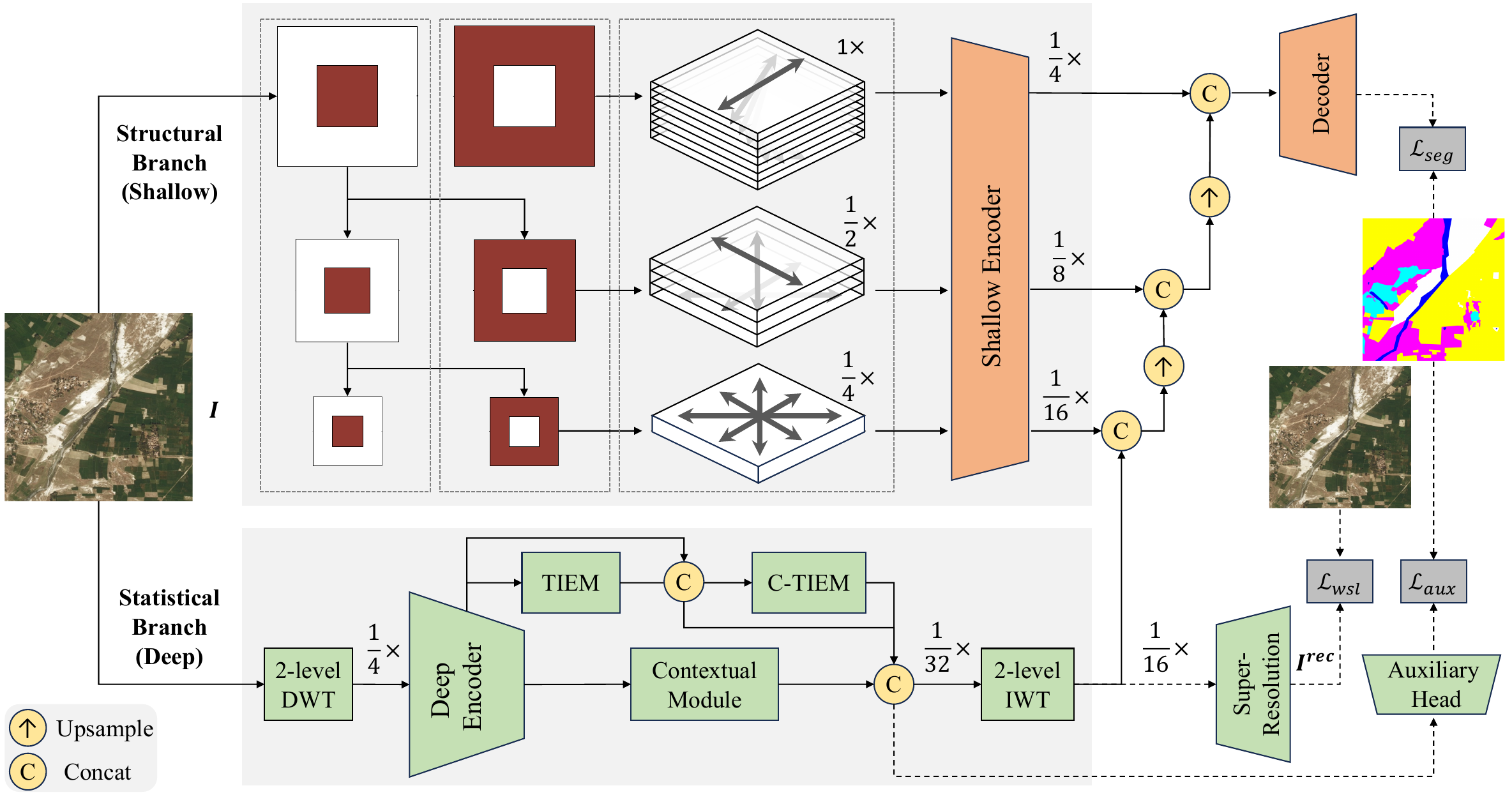}
    \caption{U-SSNet: collaborative structural and statistical texture learning design for UHR segmentation.} 
    \label{u-stlnet++}
\end{figure*}

\subsubsection{C-TIEM: Adaptation Step}

In GCLM, after producing the co-occurrence map for an image, some statistical descriptors such as contrast, uniformity, and homogeneity are employed to represent the texture information. 
Different from these hand-crafted descriptors, benefiting from the powerful feature extraction ability of deep learning, we drive the network to learn statistical representations adaptively during training. Specifically, an MLP followed by a level-wise average is employed to produce the texture feature $\mathbf{T}$ through,
\begin{equation}
\begin{split}
        &\mathbf{D}' = MLP\left(\mathbf{D}\right),\quad \mathbf{D}'\in \mathbb{R}^{C_3\times N\times N},\\
        &\mathbf{T} = \frac{\sum_{m=1}^{N}\sum_{n=1}^{N}{\mathbf{D}'_{:, m, n}}}{N \cdot N},
\end{split}
\end{equation}
\noindent where $\mathbf{T} \in \mathbb{R}^{C_3\times 1 \times 1}$, which is then upsampled to $\mathbb{R}^{C_3\times H \times W}$.

\subsection{STLNet++}

Based on C-TIEM and CDM, STLNet++ is formulated to integrate low-level structural and statistical texture learning, thereby enhancing the performance of existing semantic segmentation networks. As illustrated in Fig. \ref{stlnet++}, the base network can be any of the established classical segmentation architectures, such as DeepLabV3+ \cite{deeplabv3+}, DANet \cite{danet}, and HRNet \cite{hrnet}, which is tasked with extracting high-level contextual features. Concurrently, the structural texture feature $\mathbf{T}_{str}$ extracted by the CDM, along with the statistical texture feature $\mathbf{T}_{sta}$ derived from C-TIEM, are upsampled, combined, and subsequently fused with the contextual features to facilitate the final prediction. Notably, the proposed integration of structural and statistical texture learning operates seamlessly, synergizing with the existing high-level contextual information within the network. This collaborative approach enables a nuanced understanding and segmentation of the input images, achieving a high degree of precision in the segmentation process.

\subsection{U-SSNet}

In order to further validate the effectiveness of texture learning, we extend the proposed method to the more challenging task of ultra-high-resolution (UHR) semantic segmentation. Building upon our previous work, WSDNet \cite{urur}, we further design U-SSNet, which integrates both structural and statistical texture components. As shown in Fig. \ref{u-stlnet++}, the upper branch, known as the structural branch, employs the CDM module to generate multi-scale inputs. These inputs are then passed through a shallow encoder, extracting multi-scale features that are rich in structural texture, facilitated by a lightweight STDC \cite{stdc} network. The lower branch, known as the statistical branch, begins with a two-level Discrete Wavelet Transform (DWT) to downsample the input image, reducing the memory consumption while preserving more spatial and frequency information compared to ordinary downsampling methods. Subsequently, this downsampled input is processed by STLNet++ based on ResNet18, followed by a two-level Inverse Discrete Wavelet Transform (IWT). Finally, the outputs from the two branches are fused through multi-level upsampling and concatenation, and the final segmentation mask is then predicted by a decoder.

During training, in addition to the classic segmentation cross-entropy loss $\mathcal{L}_{seg}$, a super-resolution loss $\mathcal{L}_{wsl}$ and an auxiliary loss $\mathcal{L}_{aux}$ are added to the lower branch to enhance the training stability and improve the segmentation accuracy. Specifically, following WSDNet \cite{urur}, we employ the wavelet smooth loss as the super-resolution loss, 

\begin{equation}
\begin{aligned}
    \mathcal{L}_{wsl} = \sum_{l=1}^L \sum_{b=1}^{4^{l}}(&\lambda_1 ||I_{l,b;1} - I_{l,b;1}^{rec}||_2 + \\ &\lambda_2 \sum_{i=2}^{4}||I_{l,b;i} - I_{l,b;i}^{rec}||_1),
\end{aligned}
\end{equation}

\noindent where $I_{l,b;1}$ denotes the low-frequency subband after the $l$-th DWT, $I_{l,b;i}$ denotes the $i$-th high-frequency subband after the $l$-th DWT, $I_{l,b;1}^{rec}$ and $I_{l,b;i}^{rec}$ mean similarly, and $\lambda_1, \lambda_2$ are the weights of the low-frequency and high-frequency constrains, respectively.

\section{Experiments: Knowledge Distillation} \label{sec:exp_kd}

\subsection{Datasets and Evaluation Metrics} \label{subsec:dataset}
Extensive experiments are conducted on the following large-scale datasets to verify the effectiveness of our SSTKD.

\noindent \textbf{Cityscapes.} The Cityscapes dataset \cite{cordts2016cityscapes} is used for semantic urban scene understanding, and has 5,000 finely annotated images captured from 50 different cities, where 30 common classes are provided and 19 classes are used for evaluation and testing.

\noindent \textbf{ADE20K.}  The ADE20K dataset \cite{ade20k} is a large-scale scene parsing benchmark and composed of more than 27K images from the SUN and Places databases. Images are fully annotated with objects, spanning over 3K object categories. This dataset contains dense labels of 150 stuff/object categories, and has 20K/2K/3K images for training, validation, and testing respectively, and includes 150 classes of diverse scenes.

\noindent \textbf{Pascal VOC 2012.}  The Pascal VOC 2012 dataset \cite{everingham2010pascal} is a segmentation benchmark of 10,582/1,449/1,456 images for training, validation, and testing respectively, which involves 20 foreground object classes and one background class.

\noindent \textbf{Evaluation Metrics.}  In all the experiments, we adopt the mean Intersection-over-Union (mIoU) to study the effectiveness of distillation.
The model size is represented by the number of network parameters, and the complexity is evaluated by the sum of floating point operations (FLOPs) in one forward propagation on a fixed input size.

\subsection{Implementation Details}

Following previous works  \cite{liu2019structured,wangintra2020}, 
we adopt the segmentation architecture PSPNet   \cite{pspnet} with ResNet101   \cite{he2016deep} backbone as the teacher network, and use the PSPNet with different small backbones as the student network. More concretely, we choose the backbone of ResNet18, ResNet18$^*$, and ResNet18 (0.5) for the student network, indicating training from scratch, pretrained on ImageNet, and the width-halved version of ResNet18, respectively. For comparison with existing methods, we also replace the student backbone with EfficientNet-B0   \cite{tan2019efficientnet}, EfficientNet-B1 \cite{tan2019efficientnet}, and MobileNet \cite{MobileNetV2} that are very compact and have low complexity, aiming to validate the effectiveness when the teacher model and the student model are of different architectural types. When compared with CIRKD \cite{CIRKD} on the Cityscapes dataset, the teacher model is changed to DeepLabV3+ with ResNet101. 
In the training process of the student network, the input images are all cut into $512 \times 512$. Random scaling (from 0.5 to 2.1) and random horizontal flipping (with the probability of 0.5) are applied as the data augmentation. We implement two-level contourlet decomposition iteratively in the CDM, where $m$ is set to $4$ and $3$ respectively. We also set $N=128$ and $\theta=0.9$. Stochastic Gradient Descent with momentum is deployed as the optimizer, where the momentum is 0.9 and weight decay rate is 1e-5. The base learning rate is 0.015 and multiplied by $(1-\frac{iter}{max-iter})^{0.9}$.  We train the model for 80,000 iterations with the batch size of 16. All the models are tested under a single-scale setting.

\begin{table}[!t]
\caption{Ablation study of structural and statistical texture learning. Efficacy of both texture knowledge on Cityscapes \textit{val} set is reported. KD means knowledge distillation.}
\centering
\scalebox{0.95}{
\begin{tabular}{cc|c|c|c}
\toprule
\multicolumn{4}{c|}{Method}&\textit{val} mIoU (\%) \\ \midrule
\multicolumn{4}{c|}{T: ResNet101}& 78.56 \\
\multicolumn{4}{c|}{S: ResNet18}& 69.10 \\ \midrule
\multicolumn{2}{c|}{\begin{tabular}[c]{@{}c@{}}Response-based\\ KD\end{tabular}}&\multirow{2}{*}{\begin{tabular}[c]{@{}c@{}}Structural\\ Texture KD\end{tabular}}&\multirow{2}{*}{\begin{tabular}[c]{@{}c@{}}Statistical\\ Texture KD\end{tabular}} \\ 
\begin{tabular}[c]{@{}c@{}}output\\ prob.\end{tabular} & \begin{tabular}[c]{@{}c@{}}adver.\\ learning\end{tabular}& && \\\midrule
\checkmark&&&&70.51 (+1.41) \\
\checkmark&\checkmark&&&72.47 (+3.37) \\ \midrule
\checkmark&\checkmark&\checkmark& & 74.80 (+5.70)  \\
\checkmark&\checkmark&   & \checkmark &75.89 (+6.79) \\
\checkmark&\checkmark& \checkmark  & \checkmark &76.65 ({+7.55}) \\\bottomrule
\end{tabular}
}

\label{efficacy_of_texture}
\end{table}

\subsection{Ablation Study}

In all the ablation studies, we use the Cityscapes validation dataset, and ResNet18 pretrained on ImageNet as the backbone of the student network.

\begin{figure}[!h]
    \centering
    \includegraphics[width=\linewidth]{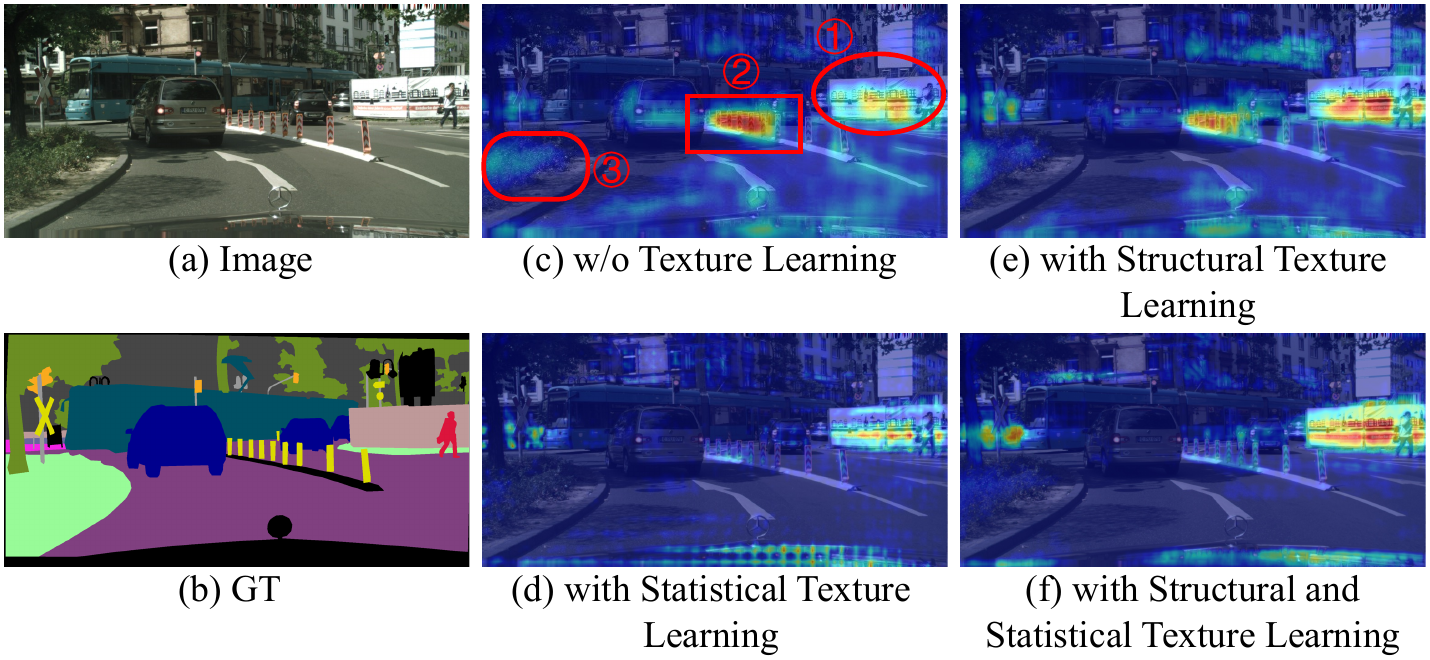}
    \caption{CAM visualization of the synergistic interaction between structural and  statistic texture learning.}
    \label{cam_texture_vis}
\end{figure}

\subsubsection{Ablation Study of Structural and Statistical Texture Knowledge Distillation}

Table \ref{efficacy_of_texture} shows the effectiveness of the two types of texture knowledge. The student network without distillation achieves an mIoU of 69.10\%, and the response knowledge improves it to 72.47\%, combining the advantages of output probability and adversarial learning. Then, we further add two kinds of texture knowledge successively to validate the effect of each one. Concretely, it shows that the structural texture knowledge brings an improvement of 5.7\%, and the statistical texture knowledge  boosts the improvement to 5.99\%. Finally, when we add both texture knowledge, the performance is promoted to 76.35\% with a large increase of 7.55\%. The gap between student and teacher is finally reduced to 1.91\%, providing a closer result to the teacher network.

\subsubsection{Synergistic Interaction between Structural and  Statistic Texture Learning}

Overall, structural texture learning tends to emphasize local features such as edges in the image, while statistical texture focuses more on global characteristics such as image contrasts. However, this does not imply that these two types of texture learning operate independently in the segmentation process. On the contrary, structural and statistical texture learning exhibit a synergistic interaction, mutually enhancing and promoting each other during the segmentation process. 
As discussed in \cite{dip, structure_dense, do2005contourlet,  kaplan1999extended}, models tend to enhance local dark regions when learning local patterns, which in turn benefits the learning of global statistical information. Conversely, when learning global patterns, models improve the delineation of local edges, facilitating the structural details. To validate this analysis, we perform a visualization on the knowledge distillation task. As shown in Fig. \ref{cam_texture_vis}, (a) and (b) depict the original image and its corresponding segmentation ground truth, (c) shows the class activation map (CAM) without texture learning, (d) and (e) display the CAMs with only statistical texture learning and structural texture learning, respectively, and (f) presents the CAM with both texture learning. Observations are as follows.

\begin{itemize}
    \item From the comparison between (c) and (e), we see that local details of  wall (region \textcircled{1}) in (c) are lost due to severe overexposure. Yet (e) significantly improves the activation range, which indicates that structural texture learning not only captures the details of the wall but also mitigates overexposure's high contrast. This demonstrates that the local learning paradigm of structural texture also enhances the global information. 
    \item From the comparison of (c) and (d), the learning failure caused by drastic light variations in regions \textcircled{1}, \textcircled{2}, and \textcircled{3} in (c) is effectively addressed in (d), where incorrect activations in regions \textcircled{2} and \textcircled{3} are significantly suppressed. Interestingly, region \textcircled{1} also shows improvement relative to  (c), confirming that the global learning paradigm of statistical texture enhances the local detail capture.
    \item Ultimately, (f) illustrates that under the synergistic learning of both structural and statistical textures, regions \textcircled{1}, \textcircled{2}, and \textcircled{3} all exhibit improved learning outcomes.
\end{itemize}

\subsubsection{Ablation Study of Structural Texture Extraction Methods}
As discussed in Sec. \ref{sec_discuss_structural}, compared to other signal representation tools, contourlet demonstrates a superior capability in capturing structural texture. In Table \ref{abl_structural}, we present an ablation study on the task of  knowledge distillation for semantic segmentation, focusing on structural extraction techniques. Specifically, we compare the FFT-based, wavelet-based, and contourlet-based (the proposed CDM) structural texture extraction methods. The results indicate that both the FFT-based and wavelet-based methods show considerable improvements over the baseline, validating the robustness and effectiveness of structural texture knowledge, regardless of the extraction techniques used. Furthermore, when employing the contourlet-based extraction method, the performance is improved by 5.70\%, experimentally confirming its significant advantage on structural texture extraction.

\begin{table}[]
\caption{Ablation study of structural texture extraction technique.}
\centering
\begin{tabular}{lc}
\toprule
Method                            &  mIoU (\%) \\ \midrule
Baseline (w/o distillation)    & 69.10    \\ \midrule
Structural Knowledge (FFT)    & 72.78    \\
Structural Knowledge (Wavelet)    & 73.13    \\
Structural Knowledge (Contourlet) & 74.80    \\ \bottomrule
\end{tabular}
\label{abl_structural}
\end{table}

\begin{table}[]
\caption{Ablation study of statistical texture extraction technique.}
\centering
\begin{tabular}{lc}
\toprule
Method                            &  mIoU (\%) \\ \midrule
Baseline (w/o distillation)    & 69.10    \\ \midrule
Statistical Knowledge (GLCM)    & 72.97    \\
Statistical Knowledge (TIEM)  & 75.89    \\ \bottomrule
\end{tabular}
\label{abl_statistical}
\end{table}

\subsubsection{Ablation Study of Statistical Texture Extraction Methods}

As discussed in Sec. \ref{sec_discuss_statistical}, the Texture Information Extraction Method (TIEM) demonstrates a superior statistical texture representation compared to existing Gray-level Co-occurrence Matrix (GLCM)-based methods. To validate this claim, we present an ablation study focusing on statistical extraction techniques in the context of knowledge distillation for semantic segmentation, as outlined in Table \ref{abl_statistical}. Specifically, we compare GLCM-based methods with our proposed TIEM statistical extraction methods. The results reveal that GLCM-based methods exhibit an improvement over the baseline, achieving an enhancement of 3.87\%. This finding supports the effectiveness of statistical texture knowledge, even when employing a relatively naive statistical texture extraction approach. Furthermore, with the application of TIEM, the performance is remarkably improved by an additional 6.79\%, thereby confirming the significant advantages of TIEM in statistical texture extraction.

\begin{table}[!t]
\centering
\caption{Parameter analysis of adaptive sampling in statistical texture learning. KD means knowledge distillation.}
\scalebox{1}{\begin{tabular}{l c}
\toprule
Method & mIoU (\%) \\ 
\midrule
Statistical Texture KD  & 75.89 \\ 
\midrule
w/o Denoising  & 75.06 \\
w/o Adaptive Sampling & 75.37 \\
Adaptive Sampling ($k=1, \beta=0.0 $) & 73.70 \\
Adaptive Sampling ($k=2, \beta=0.7$) & 75.89 \\
Adaptive Sampling ($k=8, \beta=1.0$) & 75.26 \\
\bottomrule
\end{tabular}}
\label{table_abl_sta}
\end{table}

\begin{figure}[]
\centering
\includegraphics[width=1\linewidth]{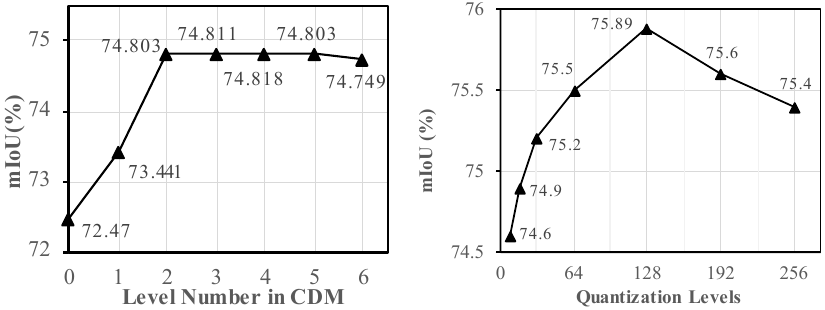}
\caption{Parameter analysis of structural and statistical texture learning. (Left) The impact of level number of contourlet decomposition, and (Right) the impact of quantization levels in the quantization step of TIEM.}
\label{layer_num}
\end{figure}

\subsubsection{Parameter Analysis of Structural Texture Learning}

We conduct experiments to verify the effectiveness of the components in the structural texture module, and show the impact of the level number of contourlet decomposition in the CDM. As shown in Fig. \ref{layer_num}, the baseline is the result of the student network with response knowledge that achieves 72.47\%, and CDM with $level=1$ achieves 73.44\%. We then increase the level number to 2, 3, 4, 5, and 6. The results show that with the level number gradually increasing, the mIoU also gradually increases and then remains around $74.8\%$, indicating that the texture knowledge extraction almost reaches saturation when the level number gets to $2$. Thus, in the main experiments, we set the level to 2 to balance the computational cost.

\begin{table}[!t]
\centering
\caption{Complexity of texture extraction modules in terms of FLOPs and parameters.}
 \centering
    \scalebox{1}{
        \begin{tabular}{c c c}
        \toprule
        Method  & Params (M)  & FLOPs (G)  \\
        \midrule
        PSPNet & 70.43 & 574.9 \\ 
        \midrule
        CDM & 1.24 & 10.90 \\
        TIEM & 2.80 & 23.73 \\
        \bottomrule
    \end{tabular}
    }
    \label{flop}
\end{table}

\begin{figure}[!h]
    \centering
    \includegraphics[width=\linewidth]{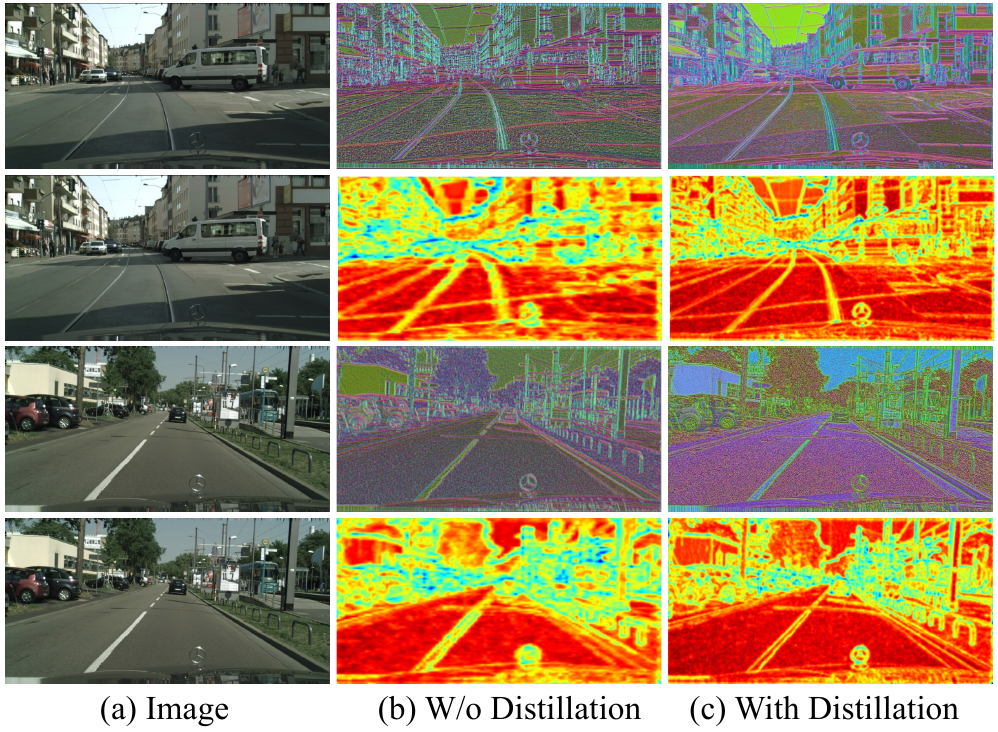}
    \caption{Comparison of visualization of the low-level feature from stage 1 of the backbone. (a) Original image, (b) results from the student network without texture knowledge distillation, and (c) changes after applying the distillation in our method. Rows 1 and 3 show the structural texture results, while rows 2 and 4 display the statistical texture results.}
    \label{texture_vis}
\end{figure}

\subsubsection{Parameter Analysis of Statistical Texture Learning}

In Table \ref{table_abl_sta}, we show the effectiveness of different components in the statistical texture module.  Overall, statistical texture knowledge distillation achieves 75.89\%. When there is no denoising step or adaptive sampler, mIoU decreases to 75.06\% and 75.37\%, respectively. Moreover, we also show the effect of parameters in the sampler. Mildly biased sampling ($k=2, \beta=0.7$) towards uncertain regions performs the best. Heavily biased sampling ($k=8, \beta=1.0$) performs even worse, indicating the importance of coverage. 
Furthermore, we also show the impact of quantization levels in the quantization step of TIEM. During the quantization process, the number of quantization levels can be chosen flexibly, and we conduct experiments to explore the most suitable choice. The results are shown in Fig. \ref{layer_num}, where the performance is unsatisfactory when the number of quantization levels is too small, and increasing the number to 128 can significantly boost mIoU to 75.89\%. However, further increasing the number may cause the performance to decrease slightly. The reason may be that a too dense quantization will result in a sparse high-dimensional feature, which may affect the training stability negatively and cause overfitting.

\subsubsection{Complexity of Texture Extraction Modules}

We show that the proposed texture modules are lightweight in Table \ref{flop}, which are estimated with the input size of $512\times 1024$. It reports that the Params and FLOPs of original PSPNet are 70.43M and 574.9G, respectively, by contrast, our CDM (1.24M, 10.90G) and TIEM (2.80M, 23.73G) only bring very little extra cost.

\begin{table}
\centering
\caption{Quantitative results on Cityscapes. ``R18" (``R101") means ResNet18 (ResNet101).} 
\scalebox{0.9}{\begin{tabular}{l|cc|c|c}
\toprule
\multirow{2}{*}{Method} & \multicolumn{2}{c|}{\begin{tabular}[c]{@{}c@{}}Cityscapes\\ mIoU (\%)\end{tabular}} & \multirow{2}{*}{\begin{tabular}[c]{@{}c@{}}Params\\ (M)\end{tabular}} & \multirow{2}{*}{\begin{tabular}[c]{@{}c@{}}Flops\\ (G)\end{tabular}} \\ \cmidrule{2-3}
                        & \multicolumn{1}{c|}{\textit{val}}  & \textit{testing}  &        &         \\ \midrule
ENet \cite{paszke2016enet}  & \multicolumn{1}{c|}{-}    & 58.3  & 0.358  & 3.612   \\ 
ICNet \cite{zhao2018icnet} & \multicolumn{1}{c|}{-}    & 69.5  & 26.50  & 28.30   \\ 
FCN \cite{fcn} & \multicolumn{1}{c|}{-}    & 62.7  & 134.5  & 333.9   \\ 
RefineNet \cite{lin2017refinenet}  & \multicolumn{1}{c|}{-}    & 73.6  & 118.1  & 525.7   \\ 
OCNet \cite{yuan2018ocnet}  & \multicolumn{1}{c|}{-}     & 80.1  & 62.58  & 548.5     \\ 
STLNet \cite{stlnet} & \multicolumn{1}{c|}{82.3} & 82.3  & 81.39  & 293.03  \\ \midrule
\midrule
\multicolumn{5}{c}{Results w/ and w/o distillation schemes} \\ \midrule
\midrule
T: PSPNet-R101 \cite{pspnet} & \multicolumn{1}{c|}{78.56} & 78.4 & 70.43& 574.9 \\ \midrule
S: PSPNet-R$18$(0.5)& \multicolumn{1}{c|}{61.17} & - & 3.271 & 31.53 \\
+ SKDS  \cite{liu2019structured}& \multicolumn{1}{c|}{61.60}& 60.50 & 3.271 & 31.53 \\
+ SKDD  \cite{structure_dense}& \multicolumn{1}{c|}{62.35}& - & 3.271 & 31.53 \\
+ IFVD  \cite{wangintra2020} & \multicolumn{1}{c|}{63.35}& 63.68 & 3.271 & 31.53 \\
+ CWD  \cite{channel_dist} & \multicolumn{1}{c|}{68.57}& 66.75 & 3.271 & 31.53 \\
+ SSTKD  & \multicolumn{1}{c|}{71.00}& 69.15 & 3.271 & 31.53 \\ \midrule
S: PSPNet-R$18$ & \multicolumn{1}{c|}{63.63} & - & 13.07 & 125.8 \\
+ SKDS  \cite{liu2019structured}& \multicolumn{1}{c|}{63.20}& 62.10 & 13.07 & 125.8 \\
+ SKDD  \cite{structure_dense}& \multicolumn{1}{c|}{64.68 }& - & 13.07 & 125.8 \\
+ IFVD  \cite{wangintra2020} & \multicolumn{1}{c|}{66.63}& 65.72 & 13.07 & 125.8 \\
+ CWD  \cite{channel_dist} & \multicolumn{1}{c|}{71.03}& 70.43 & 13.07 & 125.8 \\
+ SSTKD  & \multicolumn{1}{c|}{73.01}& 72.15 & 13.07 & 125.8 \\ \midrule
S: PSPNet-R18$^*$ & \multicolumn{1}{c|}{70.09} & 67.60 & 13.07 & 125.8 \\
+ SKDS  \cite{liu2019structured}& \multicolumn{1}{c|}{72.70}& 71.40 & 13.07 & 125.8 \\
+ SKDD  \cite{structure_dense}& \multicolumn{1}{c|}{74.08 }& - & 13.07 & 125.8 \\
+ IFVD  \cite{wangintra2020} & \multicolumn{1}{c|}{74.54}& 72.74 & 13.07 & 125.8 \\
+ CWD  \cite{channel_dist} & \multicolumn{1}{c|}{75.90}& 74.58 & 13.07 & 125.8 \\
+ SSTKD  & \multicolumn{1}{c|}{76.65} & 75.59 & 13.07 & 125.8 \\ \midrule
S: DeepLabV3-R18 & \multicolumn{1}{c|}{73.37} & 72.39 & 12.62 & 123.9 \\
+ SKDS  \cite{liu2019structured}& \multicolumn{1}{c|}{73.87}& 72.63 & 12.62 & 123.9 \\
+ IFVD  \cite{wangintra2020}& \multicolumn{1}{c|}{74.09 }& 72.97 & 12.62 & 123.9 \\
+ CWD  \cite{channel_dist} & \multicolumn{1}{c|}{75.91}& 74.32 & 12.62 & 123.9 \\
+ CIRKD  \cite{CIRKD} & \multicolumn{1}{c|}{76.40}& 75.11 & 12.62 & 123.9 \\
+ SSTKD  & \multicolumn{1}{c|}{76.69} & 75.77 & 12.62 & 123.9 \\ \midrule
S: EfficientNet-B0 & \multicolumn{1}{c|}{58.37} & 58.06 & 4.19 & 7.967 \\
+ SKDS  \cite{liu2019structured}& \multicolumn{1}{c|}{62.90}& 61.80 & 4.19 & 7.967 \\
+ IFVD  \cite{wangintra2020}& \multicolumn{1}{c|}{64.73}& 62.52 & 4.19 & 7.967 \\
+ CWD  \cite{channel_dist} & \multicolumn{1}{c|}{-}& - & 4.19 & 7.967 \\
+ SSTKD  & \multicolumn{1}{c|}{68.10} & 66.22 & 4.19 & 7.967 \\ \midrule
S: EfficientNet-B1 & \multicolumn{1}{c|}{60.40} & 59.91 & 6.70 & 9.896 \\
+ SKDS  \cite{liu2019structured}& \multicolumn{1}{c|}{63.13}& 62.59 & 6.70 & 9.896 \\
+ IFVD  \cite{wangintra2020}& \multicolumn{1}{c|}{66.50}& 64.42 & 6.70 & 9.896 \\
+ CWD  \cite{channel_dist} & \multicolumn{1}{c|}{-}& - & 6.70 & 9.896 \\
+ SSTKD  & \multicolumn{1}{c|}{69.03} & 67.08 & 6.70 & 9.896 \\ 
\bottomrule
\end{tabular}}
\label{sotacity}
\end{table}

\subsubsection{Visualization of Low-Level Feature with and without Distillation}

Fig. \ref{texture_vis} shows the visualization of the low-level feature from the student network with ResNet18 backbone. For comparison, we give the results of the student network with and without the texture knowledge distillation in our method. Specifically, Fig. \ref{texture_vis}(b) shows the results that are produced without the structural/statistical texture knowledge. As can be seen, the texture details are very fuzzy and in low-contrast, the contours of the objects are also incomplete. In contrast, they are clearer in Fig. \ref{texture_vis}(c) when incorporating the two kinds of texture knowledge. In such a case, the contours of the auto logo and lane lines can offer more correct cues for semantic segmentation. This phenomenon illustrates the validity of the texture knowledge, providing a better understanding of our method.

\begin{figure}[!h]
    \centering
    \includegraphics[width=\linewidth]{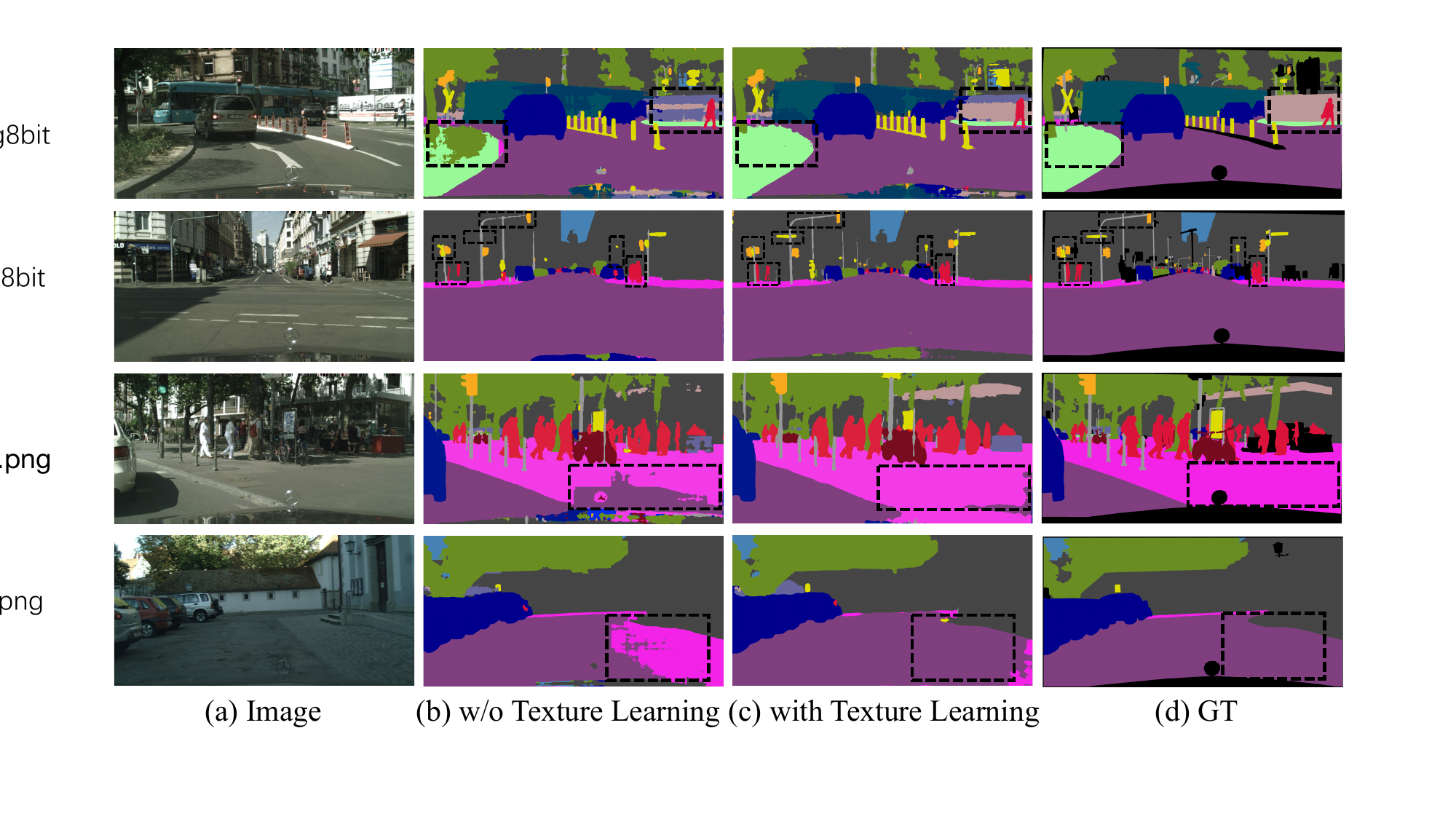}
    \caption{Visualization of the improvements in the segmentation results by the structural (rows 1 and 2) and statistical (rows 3 and 4) texture learning, respectively. }
    \label{seg_texture_vis}
\end{figure}

\subsubsection{Visualization of Improvements from Structural and Statistical Texture Learning}

Fig. \ref{seg_texture_vis} elucidates the contribution of the two types of texture learning to the final segmentation outcomes. Specifically, the first and second rows demonstrate the effects of structural texture learning, while the third and fourth rows illustrate the results of statistical texture learning. From the first and second rows, it is evident that structural texture learning significantly enhances the segmentation of local details, such as the ``grass" and ``wall" regions in the first row and the finer details of ``pedestrians", ``streetlights”, and ``signs“ in the second row. Conversely, the third and fourth rows show that statistical texture learning markedly improves the global segmentation performance. This is particularly apparent in the segmentation of the road surfaces, which indicates significant enhancement after statistical texture learning, despite the noticeable variations in illumination. These experimental results explicitly underscore the segmentation improvements brought by structural and statistical texture learning from both local and global perspectives.

\subsubsection{Visualization of Segmentation Results with and without Distillation}

Fig. \ref{city_vis} provides the visualization results for the overall effectiveness of knowledge distillation, which shows the significant enhancement of our SSTKD's segmentation ability.

\subsection{Comparison with State-of-the-Arts} 

\noindent\textbf{Segmentation.} Tables \ref{sotacity} and \ref{sota_voc_ade} show that the proposed SSTKD framework achieves state-of-the-art results with different backbones on Cityscapes, PASCAL VOC 2012, and ADE20K datasets, respectively. More comprehensively, SSTKD improves the student model (PSPNet) built on ResNet18 (pretrained by scratch) to 73.01\% and 72.15\% on the validation and testing datasets, respectively, while a channel-halved variant of ResNet18 (0.5) backbone shows a consistent improvement. When changing the student backbone to ResNet18 pretrained on ImageNet, the performance is further boosted to 76.65\% and 75.59\%. Moreover, we change the student backbone to Deeplab-R18, EfficientNet-B0, and EfficientNet-B1 to demonstrate the generality of our SSTKD by testing the robustness when the student network is of different architectural types from the teacher network. 
In such cases, our method also obtains consistent improvements, e.g., deeplab-R18 student improves 3.32\% and 3.38\%, higher than the baseline model on the validation and testing sets. When changing to EfficientNet-B0 and EfficientNet-B1, the gains are 9.73\%/8.16\% and 8.63\%/7.17\%, respectively, which are also larger than the previous methods. In conclusion, our method exceeds all existing works with different backbones and network architectures. 

\noindent\textbf{Detection. } To further validate the generalizability, versatility, and effectiveness of SSTKD in dense prediction tasks, we conduct additional evaluations in the context of object detection. Specifically, we follow the well-known approach outlined in SKDD \cite{structure_dense}, employing the one-stage FCOS \cite{fcos} detection network for both the teacher and student models. The teacher model utilizes a ResNext-101-FPN \cite{resnext} backbone, while the student model is based on MobileNet V2 \cite{MobileNetV2}. For training, we adhere to the training schedule established by SKDD, conducting approximately 90K iterations. Table \ref{sota_det} presents a comparative analysis of the distillation effects of SSTKD against SKDD and another classical method, MIMIC \cite{mimic}. The results demonstrate that SSTKD exhibits significant performance advantages across all metrics, thereby substantiating the  effectiveness of our proposed structural and statistical texture learning-based knowledge distillation method in dense prediction tasks.

\begin{figure}[!t]
    \centering
    \includegraphics[width=1\linewidth]{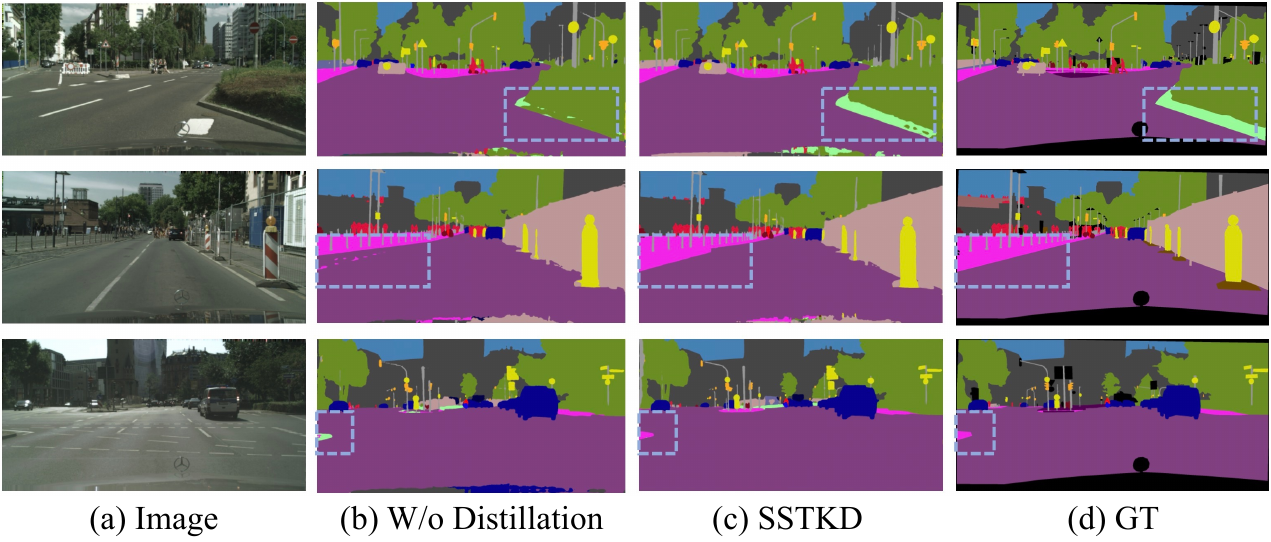} 
    \caption{Visualization of segmentation results with and without distillation on Cityscapes dataset. SSTKD improves the student network w/o distillation to produce more accurate and detailed results that are circled by dotted lines.}
    \label{city_vis}
\end{figure}

\begin{table}
\centering
\caption{Quantitative results of knowledge distillation for object detection on COCO-minival dataset.}
\scalebox{0.9}{\begin{tabular}{l|ccc}
\toprule
Method  & mAP (\%)  & AP50 (\%) & AP75 (\%) \\ \midrule
Teacher (FCOS, ResNeXt-101-FPN) & 42.5 & 61.7 & 45.9 \\ 
Student (FCOS, MobileNetV2) & 31.0 & 48.5 & 32.7 \\ \midrule
+MIMIC \cite{mimic}  & 31.4 & 48.1 & 33.4 \\
+SKDD \cite{structure_dense}   & 32.1 & 49.5 & 34.2 \\
+SSTKD  & 32.8 & 49.9 & 35.1 \\ \bottomrule
\end{tabular}}
\label{sota_det}
\end{table}

\begin{table}[]
\centering
\caption{Quantitative results on Pascal VOC 2012 and ADE20K. ``R18" (``R101") means ResNet18 (ResNet101)}. 
\scalebox{0.9}{\begin{tabular}{lccc}
\toprule
\multicolumn{1}{l|}{Method}                                             & \multicolumn{1}{c|}{\begin{tabular}[c]{@{}c@{}}Pascal VOC \\ mIoU (\%)\end{tabular}} & \multicolumn{1}{c|}{\begin{tabular}[c]{@{}c@{}}ADE20K\\ mIoU (\%)\end{tabular}} & \begin{tabular}[c]{@{}c@{}}Params \\ (M)\end{tabular} \\ \midrule
\multicolumn{1}{l|}{FCN \cite{fcn}}          & \multicolumn{1}{c|}{69.6}                                                           & \multicolumn{1}{c|}{39.91}                                                     & 134.5                                                 \\
\multicolumn{1}{l|}{RefineNet \cite{lin2017refinenet}} & \multicolumn{1}{c|}{82.4}                                                           & \multicolumn{1}{c|}{40.7}                                                      & 118.1                                                 \\
\multicolumn{1}{l|}{Deeplab V3 \cite{chen2017deeplab}} & \multicolumn{1}{c|}{77.9}                                                           & \multicolumn{1}{c|}{44.99}                                                     & 87.1                                                  \\
\multicolumn{1}{l|}{PSANet \cite{psanet}}              & \multicolumn{1}{c|}{77.9}                                                           & \multicolumn{1}{c|}{43.47}                                                     & 78.13                                                 \\
\multicolumn{1}{l|}{OCRNet \cite{ocrnet}}              & \multicolumn{1}{c|}{80.3}                                                           & \multicolumn{1}{c|}{43.7}                                                      & 70.37                                                 \\ \midrule \midrule
\multicolumn{4}{c}{Results w/ and w/o distillation schemes}                                                                                                                                                                                                                                            \\ \midrule \midrule
\multicolumn{1}{l|}{T: PSPNet-R101 \cite{pspnet}}      & \multicolumn{1}{c|}{78.52}                                                          & \multicolumn{1}{c|}{44.39}                                                     & 70.43                                                 \\ \midrule
\multicolumn{1}{l|}{S: PSPNet-R18}                                      & \multicolumn{1}{c|}{65.42}                                                          & \multicolumn{1}{c|}{24.65}                                                     & 13.07                                                 \\
\multicolumn{1}{l|}{+SKDS \cite{liu2019structured}}    & \multicolumn{1}{c|}{67.73}                                                          & \multicolumn{1}{c|}{25.11}                                                     & 13.07                                                 \\
\multicolumn{1}{l|}{+IFVD \cite{wangintra2020}}        & \multicolumn{1}{c|}{68.04}                                                          & \multicolumn{1}{c|}{25.72}                                                     & 13.07                                                 \\
\multicolumn{1}{l|}{+CWD \cite{structure_dense}}      & \multicolumn{1}{c|}{69.25}                                                          & \multicolumn{1}{c|}{26.80}                                                     & 13.07                                                 \\
\multicolumn{1}{l|}{+SSTKD}                                             & \multicolumn{1}{c|}{70.98}                                                          & \multicolumn{1}{c|}{29.19}                                                     & 13.07                                                 \\ \midrule
\multicolumn{1}{l|}{S: PSPNet-MBV2}                                     & \multicolumn{1}{c|}{62.38}                                                          & \multicolumn{1}{c|}{23.15}                                                     & 1.98                                                  \\
\multicolumn{1}{l|}{+SKDS \cite{liu2019structured}}    & \multicolumn{1}{c|}{63.95}                                                          & \multicolumn{1}{c|}{24.79}                                                     & 1.98                                                  \\
\multicolumn{1}{l|}{+IFVD \cite{wangintra2020}}        & \multicolumn{1}{c|}{64.73}                                                          & \multicolumn{1}{c|}{25.33}                                                     & 1.98                                                  \\
\multicolumn{1}{l|}{+CWD \cite{structure_dense}}      & \multicolumn{1}{c|}{65.93}                                                          & \multicolumn{1}{c|}{27.97}                                                     & 1.98                                                  \\
\multicolumn{1}{l|}{+SSTKD}                                             & \multicolumn{1}{c|}{66.35}                                                          & \multicolumn{1}{c|}{29.02}                                                     & 1.98                                                  \\ \midrule
\multicolumn{1}{l|}{S: Deeplab-R18}                                     & \multicolumn{1}{c|}{66.81}                                                          & \multicolumn{1}{c|}{24.89}                                                     & 12.62                                                 \\
\multicolumn{1}{l|}{+SKDS \cite{liu2019structured}}    & \multicolumn{1}{c|}{68.13}                                                          & \multicolumn{1}{c|}{25.52}                                                     & 12.62                                                 \\
\multicolumn{1}{l|}{+IFVD \cite{wangintra2020}}        & \multicolumn{1}{c|}{68.42}                                                          & \multicolumn{1}{c|}{26.53}                                                     & 12.62                                                 \\
\multicolumn{1}{l|}{+CWD \cite{structure_dense}}      & \multicolumn{1}{c|}{69.97}                                                          & \multicolumn{1}{c|}{27.37}                                                     & 12.62                                                 \\
\multicolumn{1}{l|}{+SSTKD}                                             & \multicolumn{1}{c|}{71.45}                                                          & \multicolumn{1}{c|}{29.79}                                                     & 12.62                                                 \\ \midrule
\multicolumn{1}{l|}{S: Deeplab-MBV2}                                    & \multicolumn{1}{c|}{50.80}                                                          & \multicolumn{1}{c|}{24.98}                                                     & 2.45                                                  \\
\multicolumn{1}{l|}{+SKDS \cite{liu2019structured}}    & \multicolumn{1}{c|}{52.11}                                                          & \multicolumn{1}{c|}{26.10}                                                     & 2.45                                                  \\
\multicolumn{1}{l|}{+IFVD \cite{wangintra2020}}        & \multicolumn{1}{c|}{53.39}                                                          & \multicolumn{1}{c|}{27.25}                                                     & 2.45                                                  \\
\multicolumn{1}{l|}{+CWD \cite{structure_dense}}      & \multicolumn{1}{c|}{54.62}                                                          & \multicolumn{1}{c|}{29.18}                                                     & 2.45                                                  \\
\multicolumn{1}{l|}{+SSTKD}                                             & \multicolumn{1}{c|}{56.21}                                                          & \multicolumn{1}{c|}{31.92}                                                     & 2.45                                                  \\ \bottomrule
\end{tabular}}
\label{sota_voc_ade}
\end{table}

\section{Experiments: Semantic Segmentation}  \label{sec:exp_seg}

\subsection{Datasets and Evaluation Metrics}

We conduct experiments on Cityscapes, Pascal Context \cite{everingham2010pascal}, and ADE20K \cite{ade20k} datasets to verify the effectiveness of our STLNet++. The U-SSNet is verified on DeepGlobe \cite{deepglobe}, Inria Aerial \cite{inria}, and URUR \cite{urur} datasets. In addition to the dataset description in Sec. \ref{subsec:dataset}, we introduce the remaining datasets below, including Pascal Context, DeepGlobe, Inria Aerial, and URUR.

\noindent \textbf{Pascal Context.}  The PASCAl Context \cite{everingham2010pascal} is an extended dataset for PASCAL 2010 by providing annotations for the whole scene. It contains 4,998 images for training and 5,105 images for validation.

\noindent \textbf{DeepGlobe.} The DeepGlobe \cite{deepglobe} dataset has 803 UHR images. Each image contains $2,448 \times 2,448$ pixels and the dense annotation contains seven classes of landscape regions.
Following  \cite{glnet,fctl}, we split images into training, validation, and testing sets with 455, 207, and 142 images, respectively.

\noindent \textbf{Inria Aerial.}  The Inria Aerial \cite{inria} Challenge dataset has 180 UHR images captured from five cities. Each image contains $5,000 \times 5,000$ pixels and is annotated with a binary mask for building/non-building areas. 
We follow the protocols in \cite{glnet,fctl} by splitting images into training, validation, and testing sets with 126, 27, and 27 images, respectively.

\noindent \textbf{URUR.}  The URUR \cite{urur} is the most challenging UHR segmentation dataset to date. The dataset has 3,008 UHR images captured from 63 cities. Each image contains $5,012 \times 5,012$ pixels and is annotated with 8 fine-grained classes.

\noindent \textbf{Evaluation Metrics.}  The same metrics are used as in Sec. \ref{subsec:dataset}. F1 score is also utilized when evaluating the U-SSNet.

\subsection{Implementation Details}

For STLNet++, we follow  \cite{stlnet} and set the quantization level to 128 in 1D QCO and 8 in 2D QCO, respectively. For fair comparison, stochastic gradient descent (SGD) is used as the optimizer. For training, we use ``poly" policy to set the learning rate. We apply random scaling in the range
of [0.5, 2], random cropping, and random left-right flipping for data augmentation. For Cityscapes, we set the initial learning rate as 0.01, weight decay as 0.0005, crop size as 769×769, batch size as 8, and training iteration as 40K. For PASCAL Context, we set the initial learning rate as 0.001,
weight decay as 0.0001, crop size as 513×513, batch size as 16, and training iterations as 30K. For ADE20K, we set the initial learning rate as 0.01, weight decay as 0.0005, crop size as 513 × 513, batch size as 16, and training iteration as 100K. For U-SSNet, we follow \cite{urur} and adopt the same training settings for fair comparison. Memory and Frames-per-second (FPS) are measured on a RTX 2080Ti GPU with a batch size of 1.

\begin{table}[]
\centering
\caption{Results comparison on Cityscapes, Pascal Context, and ADE20K \textit{val} sets.}
\scalebox{0.82}{\begin{tabular}{c|c|cccc}
\toprule
Method                                           & Backbone   & \begin{tabular}[c]{@{}c@{}}Cityscapes\\ mIoU (\%)\end{tabular} & \begin{tabular}[c]{@{}c@{}}Pascal\\ Context\\ mIoU (\%)\end{tabular} & \begin{tabular}[c]{@{}c@{}}ADE20K\\ mIoU (\%)\end{tabular} & \begin{tabular}[c]{@{}c@{}}Param. \\ (M) \end{tabular} \\ \midrule
DeepLabV3+ \cite{deeplabv3+}    & ResNet101 & 79.20                                                         & 48.26                                                               & 46.35                                                     & 79            \\
CPNet \cite{cpnet}              & ResNet101 & 81.30                                                         & 53.9                                                                & 46.27                                                     & -             \\
RecoNet \cite{reconet}                                & ResNet101 & -                                                             & 55.4                                                                & 45.54                                                     & -             \\
CCNet \cite{ccnet}              & ResNet101 & 81.30                                                         & -                                                                   & 45.22                                                     & -             \\
OCRNet \cite{ocrnet}            & ResNet101 & 82.33                                                         & -                                                                   & 44.90                                                     & 80            \\
SETR \cite{zheng2021rethinking} & ViT-L/16   & 82.15                                                         & 55.8                                                                & 50.28                                                     & 318           \\
Segformer \cite{segformer}      & ViT-L/16   & 81.30                                                         & -                                                                   & 48.81                                                     & 85            \\
MaskFormer \cite{maskformer}    & Swin-B     & 82.57                                                         & -                                                                   & 48.80                                                     & 102           \\
Swin  \cite{swin}                         & Swin-B     & -                                                             & -                                                                   & 50.76                                                     & 88            \\
Segmenter \cite{segmenter}      & ViT-B/16   & 80.60                                                         & 53.9                                                                & 48.05                                                     & 103           \\
Segmenter \cite{segmenter}      & ViT-L/16   & 81.30                                                         & 56.5                                                                & 52.25                                                     & 337           \\
RegProxy \cite{RegProxy}        & ViT-B/16   & 82.20                                                         & 55.2                                                                & 51.70                                                     & 88            \\ \midrule
STLNet \cite{stlnet}            & ResNet101 & 82.30                                                         & 55.8                                                                & 46.88                                                     & 80            \\
\begin{tabular}[c]{@{}c@{}}STLNet++ \\ (w/o \textit{structural}) \end{tabular}    & ResNet101 & \textbf{83.41}                                                & \textbf{56.9}                                                       & 48.81                                                     & 81 \\
\begin{tabular}[c]{@{}c@{}}STLNet++ \\ (with \textit{structural}) \end{tabular}      & ResNet101 & \textbf{83.76}                                                & \textbf{57.4}                                                       & 49.20                                                     & 82 
\\ \bottomrule
\end{tabular}}
\label{stlnet++:cityscapes}
\end{table}

\begin{figure}[!h]
    \centering
    \includegraphics[width=0.8\linewidth]{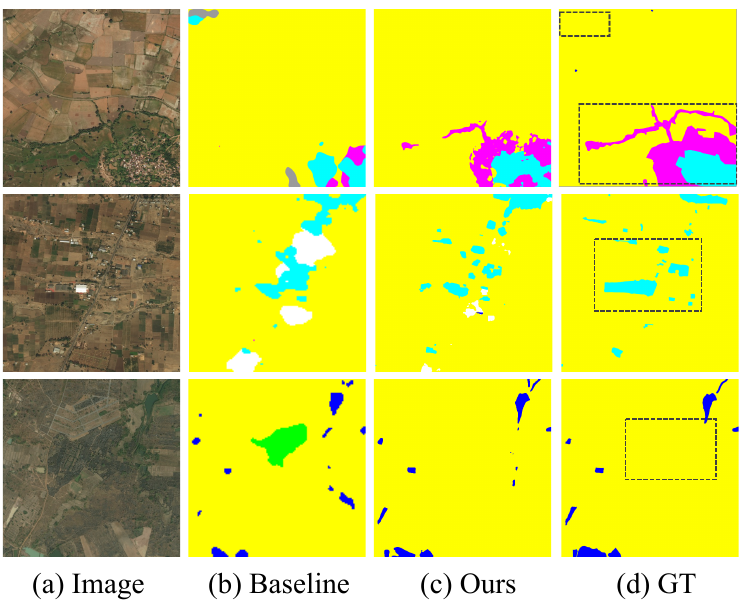}
    \caption{Comparison of visualization on DeepGlobe dataset.}
    \label{img:vis_deepglobe}
\end{figure}

\subsection{Generic Segmentation}

The comparison of STLNet++ with state-of-the-art methods on Cityscapes, Pascal Context, and ADE20K \textit{val} sets is reported in Table \ref{stlnet++:cityscapes}.
Especially, we track the research hotspots of Vison Transformer (ViT) for semantic segmentation and present the results of main latest representative works, including SETR \cite{zheng2021rethinking}, Segformer \cite{segformer}, Swin \cite{swin}, MaskFormer \cite{maskformer}, Segmenter \cite{segmenter}, and RegProxy \cite{RegProxy}.
As shown in Table \ref{stlnet++:cityscapes}, it is noteworthy that the ViTs with lightweight backbones (including ViT-B/16, Swin-T) contain the similar number of parameters as  CNN networks, while larger ones (such as ViT-L/16, Swin-B) contain tremendous parameters. However, STLNet++ also achieves remarkable performance and outperforms all other models on both Cityscapes and Pascal Context datasets, even the large ViTs, showing the charm of low-level texture information. Moreover, STLNet++ achieves the comparable performance with lightweight ViTs on ADE20K dataset, the validity is further proved. Meanwhile, we notice that there is still a performance gap between STLNet++ and some ViTs with large backbones, such as Segmenter (ViT-L/16) and HRViT-b3, on ADE20K dataset. Reasons for promising results of the latter are as follows. (1) Efficacy of tremendous parameters and enormous computation cost. ViT-L/16 has even 4 times more parameters than ResNet101. (2) Specially customized training techniques for ViTs following  \cite{trainvit}, including more large-scale pretraining on ImageNet-21K \cite{imagenet21k}, collection of robust regularization methods, and combination of data augmentation techniques such as Mixup \cite{mixup} and RandAugment \cite{randaugment}.

\begin{table}[t]
    \centering
    \caption{Comparison with state-of-the-arts on DeepGlobe \textit{testing} set.}
    \scalebox{0.9}{\begin{tabular}{l | c c c c c}
      \toprule[0.8pt]
      \textbf{Generic Models} & mIoU (\%) & F1 (\%) &  Acc (\%) & Mem (M) \\ \midrule
    \textbf{\textit{Local Inference}} & & & & \\
    U-Net \cite{unet} & 37.3  & -   & -  &  949 \\
    ICNet \cite{zhao2018icnet}  & 35.5  & - & - & 1195   \\
    PSPNet \cite{pspnet} & 53.3 & - & - & 1513 \\
    DeepLabv3+ \cite{deeplabv3+} & 63.1 & -  & - & 1279   \\
    FCN-8s \cite{fcn} & 71.8  & 82.6 & 87.6 & 1963 \\
    GLNet$^*$ \cite{glnet} & 57.3 & 64.6 & 72.7 & 1189 \\ \midrule
    
    \textbf{\textit{Global Inference}} & & & &  \\
    U-Net \cite{unet} & 38.4  & -   & -  & 5507  \\
    ICNet \cite{zhao2018icnet}  & 40.2  & - & - & 2557   \\
    PSPNet \cite{pspnet} & 56.6 & - & - & 6289  \\
    DeepLabv3+ \cite{deeplabv3+} & 63.5 & -  & - & 3199  \\
    FCN-8s \cite{fcn} & 68.8  & 79.8 & 86.2 & 5227  \\
    GLNet$^*$ \cite{glnet} & 66.4 & 79.5 & 85.8 & 1189 \\ \toprule
    
    \textbf{UHR Models} &  &  &    &  \\ \midrule
    CascadePSP \cite{cascadepsp} & 68.5 & 79.7 & 85.6 & 3236  \\
    GLNet \cite{glnet}  & 71.6 & 83.2 & 88.0  & 1865 \\
    ISDNet \cite{isdnet} & 73.3 & - & -  & 1948 \\
    FCtL \cite{fctl}  & 73.5 & 83.8 & 88.3 & 3167  \\ 
    WSDNet \cite{urur} & 74.1 & 85.2 & 89.1 & 1876 \\
    U-SSNet     & \textbf{76.9} & \textbf{86.2} & \textbf{90.1} & 1987 \\ 
     
    \bottomrule
    \end{tabular}}
    \label{ustlnet++:deepglobe}
\end{table}

\begin{table}[]
\centering
\caption{Comparison with state-of-the-arts on Inria Aerial and URUR \textit{testing} sets.}
\scalebox{0.9}{\begin{tabular}{lcccc}
\toprule
\multicolumn{1}{l|}{\multirow{2}{*}{\textbf{Generic Models}}} & \multicolumn{2}{c|}{Inria Aerial}           & \multicolumn{2}{c}{URUR} \\
\multicolumn{1}{l|}{}                                         & mIoU (\%) & \multicolumn{1}{c|}{Mem (M)} & mIoU (\%)  & Mem (M)  \\ \midrule
\multicolumn{1}{l|}{DeepLabV3+\cite{deeplabv3+}}                               & 55.9      & \multicolumn{1}{c|}{5122}       & 33.1       & 5508        \\
\multicolumn{1}{l|}{STDC\cite{stdc}}                                     & 72.4      & \multicolumn{1}{c|}{7410}       & 42.0       & 7617        \\ \midrule
\multicolumn{5}{l}{\textbf{UHR Models}}                                                                                                \\ \midrule
\multicolumn{1}{l|}{GLNet\cite{glnet}}                                    & 71.2      & \multicolumn{1}{c|}{2663}       & 41.2       & 3063        \\
\multicolumn{1}{l|}{FCtL\cite{fctl}}                                     & 73.7      & \multicolumn{1}{c|}{4332}       & 43.1       & 4508        \\
\multicolumn{1}{l|}{ISDNet\cite{isdnet}}                                   & 74.2      & \multicolumn{1}{c|}{4680}       & 45.8       & 4920        \\
\multicolumn{1}{l|}{WSDNet\cite{urur}}                                   & 75.2      & \multicolumn{1}{c|}{4379}       & 46.9       & 4560        \\
\multicolumn{1}{l|}{U-SSNet}                                  & \textbf{77.5}      & \multicolumn{1}{c|}{4502}       & \textbf{48.0}       & 4643        \\ \bottomrule
\end{tabular}}
\label{ustlnet++:inria}
\end{table}

\subsection{UHR Segmentation}

Tables \ref{ustlnet++:deepglobe} and \ref{ustlnet++:inria} show the comparisons of U-SSNet with state-of-the-arts on DeepGlobe, Inria Aerial and URUR \textit{testing} sets, respectively. Since most of these methods are not designed for ultra-high-resolution images, we follow  \cite{fctl} and denote them as ``Generic Model". Generally, two ways can be chosen for training these models on ultra-high-resolution images: (1) training with the downsampled global images, and (2) training with the local cropped patches and merging their results. The former can preserve global contextual information but may lose local details, and the latter is just the opposite. We provide the results of the two ways denoted as ``Global Inference" and ``Local Inference", respectively. On the contrary, we denote the rest models as ``Ultra-High-Resolution(UHR) Model", which are designed for ultra-high-resolution images. The original paper of GLNet \cite{glnet} provides the results for ``Local Inference" and ``Global Inference" without its global-local feature sharing module, and we denote them as GLNet$^*$. The experiments show that our method outperforms existing state-of-the-art methods on all the three metrics. More comprehensively, we outperform GLNet and FCtL by a large margin on mIoU, which can directly show the segmentation effectiveness and performance improvement. Moreover, the categories in the dataset are often distributed with serious imbalance, so we exploit the F1 score and accuracy metrics to reflect the improvements. Experimental results show that the proposed method also achieves the highest scores among all the models. 
We also report the comparison of memory usage, showing that U-SSNet is applicable for practical scenes and occupies less GPU memory. Visualization of comparisons on DeepGlobe dataset is shown in Fig. \ref{img:vis_deepglobe}.

\section{Conclusion} \label{sec:conclude}

In this paper, we focus on the low-level structural and statistical knowledge in distillation for semantic segmentation. Specifically, we propose the Contourlet Decomposition Module to effectively extract the structural texture knowledge, and the Texture Intensity Equalization Module to describe and enhance the statistical texture knowledge, respectively \cite{sstkd}. Under different supervisions, we force the student network to mimic the teacher network better from a broader perspective. Moreover, we develop the C-TIEM and generic STLNet++ to enable existing segmentation networks to harvest the low-level statistical texture information more effectively. Comprehensive experimental results demonstrate the state-of-the-art performance of the proposed methods on three segmentation tasks with seven popular benchmark datasets.

\noindent\textbf{Application.} The proposed SSTKD framework is a versatile knowledge distillation approach that extends beyond the segmentation tasks primarily addressed in this study. It is applicable to a wide range of other dense prediction tasks (object detection, depth estimation, and so on). Experimental results in Table \ref{sota_det} demonstrate SSTKD's superiority over existing distillation methods for detection tasks, indicating its effectiveness in leveraging knowledge from teacher models to enhance the student model performance. This capability highlights SSTKD's potential to advance dense prediction tasks, making it a valuable resource for researchers and practitioners. Additionally, its adaptability to various architecture positions make SSTKD a promising direction for future research in knowledge distillation.

\noindent\textbf{Novelty.} Existing knowledge distillation methods for dense prediction primarily focus on high-level features, employing various sophisticated techniques to harvest global context and long-range relational dependencies among pixels. In contrast, our SSTKD introduces low-level texture, which is crucial for characterizing images, into the realm of knowledge distillation for dense prediction for the first time. This approach innovatively proposes structural and statistical texture learning paradigms, along with the corresponding CDM and TIEM modules. Extensive experiments demonstrate that SSTKD exhibits significant performance advantages and generalizability in both segmentation and detection tasks, which are critical for dense prediction. Furthermore, we validate the efficacy of this texture learning paradigm in extensive applications including ultra-high-resolution image segmentation tasks, where it shows promising results.

\section{Potential Limitation and Broader Impact} \label{sec:outlook}

\noindent\textbf{Potential Limitation. }
The statistical texture learning approaches demonstrate higher efficacy when integrated with CNNs. However, they may need some further sophisticated designs when combined with the most recent ViT architectures, Mamba networks \cite{mamba}, or large models \cite{liu2023llava,ji2024tree}, since these architectures designed for segmentation typically crop the input images into multiple very small patches to generate the input embedding sequence, which leads to a frustrating fracture of texture, thus compromising the potential of low-level texture characterization.
In addition, although UHR segmentation methods based on texture learning surpass current state-of-the-art methods, they may also encounter some memory usage issues as image resolution continues to increase. Addressing such memory constraints may involve more granular patch cropping of the original images to reduce memory consumption.

\noindent\textbf{Broader Impact. }
Our method demonstrates broad applicability, especially in scenarios characterized by complex and deceptive textures. On the one hand, the proposed texture learning method is essentially a generalized approach to integrate basic texture features of images with deep neural networks, and it can be widely applied to various dense prediction tasks such as depth estimation and object detection. On the other hand, the proposed texture distillation method represents a foundational approach to low-level distillation, which can be seamlessly integrated with existing high-level distillation techniques, providing new avenues for the research community to explore in the realm of low-level distillation research.

\ifCLASSOPTIONcaptionsoff
  \newpage
\fi

\bibliographystyle{ieeetr}

\bibliography{bibtex}

\end{document}